%% file: main.tex
\documentclass[letterpaper,twocolumn,10pt]{article}
\usepackage{usenix2019_v3}

\usepackage{tikz}
\usepackage{amsmath}

\usepackage{filecontents}

\usepackage[numbers,sort]{natbib}

\usepackage{enumitem}
\usepackage{graphicx}
\usepackage{comment}
\usepackage{amsmath}
\usepackage{float}
\usepackage{algorithm}
\usepackage{algorithmic}
\usepackage{tikz}
\usepackage{hyperref}  
\usepackage{comment}
\usepackage{multirow}
\usepackage{booktabs}
\usepackage{setspace}
\usepackage[normalem]{ulem}

\newcommand*\circled[1]{\tikz[baseline=(char.base)]{
            \node[circle,fill=.,inner sep=0.8pt] (char) {\textcolor{white}{#1}};}}

\usepackage{color}
\definecolor{codegreen}{rgb}{0,0.6,0}
\definecolor{codegray}{rgb}{0.5,0.5,0.5}
\definecolor{codepurple}{rgb}{0.58,0,0.82}
\definecolor{backcolour}{rgb}{0.95,0.95,0.92}
\definecolor{textblue}{rgb}{.2,.2,.7}
\definecolor{textred}{rgb}{0.54,0,0}
\definecolor{textgreen}{rgb}{0,0.43,0}
\definecolor{codered}{rgb}{201,72,12}

\usepackage[T1]{fontenc}
\usepackage[scaled=0.85]{beramono} 
\usepackage{listings}
\usepackage{multirow}
\usepackage{amsfonts}

\usepackage{caption}
\usepackage{subcaption}
\usepackage[available, functional, reproduced]{usenixbadges}
 
\def\Snospace~{\S{}}

\newcommand{\boyuan}[1]{{\color{red}[Boyuan: #1]}}

\usepackage{arydshln}

\begin{document}

\date{}

\title{\Large \bf Faith: An Efficient Framework for Transformer Verification on GPUs}


\author{Boyuan Feng, Tianqi Tang, Yuke Wang, Zhaodong Chen, Zheng Wang, Shu Yang, \\ Yuan Xie, and Yufei Ding\\
University of California, Santa Barbara \\
\{boyuan,tianqi\_tang,yuke\_wang, chenzd15thu, zheng\_wang, shuyang1995, \\ yuanxie, yufeiding\}@ucsb.edu
}

\maketitle

\pagenumbering{gobble}

\input{text/0_Abstract}
\input{text/1_Introduction}

\input{text/2_Background}

\input{text/3_GraphOptimization}

\input{text/4_KernelOptimization}

\input{text/5_ExpertGuidedAutotuning}

\input{text/6_Evaluation}

\input{text/7_Discussion}

\input{text/8_Concolusion}

\vspace{-10pt}
\section{Acknowledgements}
\vspace{-7pt}

We would like to thank the anonymous reviewers and the shepherd.
This work was supported in part by NSF 2124039.

\bibliographystyle{plain}
\bibliography{reference}

\end{document}

%% file: text/0_Abstract.tex
\begin{abstract}
Transformer verification draws increasing attention in machine learning research and industry. It formally verifies the robustness of transformers against adversarial attacks such as exchanging words in a sentence with synonyms. However, the performance of transformer verification is still not satisfactory due to bound-centric computation which is significantly different from standard neural networks. In this paper, we propose \textbf{Faith}\footnote{The project is open-sourced at https://github.com/BoyuanFeng/Faith}, an efficient framework for transformer verification on GPUs. We first propose a semantic-aware computation graph transformation to identify semantic information such as bound computation in transformer verification. We exploit such semantic information to enable efficient kernel fusion at the computation graph level. Second, we propose a verification-specialized kernel crafter to efficiently map transformer verification to modern GPUs. This crafter exploits a set of GPU hardware supports to accelerate verification-specialized operations which are usually memory-intensive. Third, we propose an expert-guided autotuning to incorporate expert knowledge on GPU backends to facilitate large search space exploration. Extensive evaluations show that Faith achieves $2.1\times$ to $3.4\times$ ($2.6\times$ on average) speedup over state-of-the-art frameworks.
\end{abstract}

%% file: text/1_Introduction.tex
\vspace{-5pt}
\section{Introduction}
\vspace{-5pt}

Transformers \cite{transformer,BERT,xlnet,roberta,GPT,RaffelSRLNMZLL20,GShard} is an important category of neural networks (NNs) in machine learning research and industry.
Transformers are first designed for natural language processing (NLP) and have achieved state-of-the-art accuracy across many NLP tasks such as neural machine translation \cite{NMT1,NMT2,NMT3} and sentiment analysis \cite{sentiment1,sentiment2,sentiment3}.
Due to its success, transformers have been widely used in many industrial products such as Facebook for hate speech detection \cite{facebook} and Alexa for question answering \cite{alexa}.
Recently, transformers also show extraordinary accuracy for many computer vision tasks \cite{DosovitskiyB0WZ21,DBLP:conf/cvpr/KimLKKK21,DBLP:conf/iclr/ZhuSLLWD21,ye2019cross,DBLP:conf/cvpr/YangYFLG20} and become the new trending model.
However, similar to prior NNs, transformers are also vulnerable to adversarial attacks that add imperceptible perturbations to input data for maliciously changing transformer predictions \cite{attack1,attack2,attack3,attack4,attack5}.
One specific example of adversarial attack is to exchange words (\textit{e.g.}, cold) in a sentence with carefully selected synonyms (\textit{e.g.}, frigid).
This vulnerability may result in security concerns for real-world applications.
For example, an intentionally crafted hate speech may spread widely on social network.

\begin{figure}
    \centering
    \includegraphics{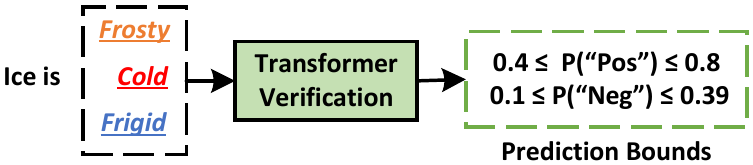}
    \vspace{-5pt}
    \caption{Illustration of transformer verification. Here, all perturbed inputs share the same prediction ``positive'' since the lower bound probability for ``positive'' (0.4)  is higher than the upper bound probability for ``negative'' (0.39).}
    \label{fig:intro_example}
    \vspace{-10pt}
\end{figure}

Transformer verification has been proposed to formally verify the robustness of a transformer against adversarial attacks \cite{reluplex,transformer_verification,transformer_verification1,DBLP:conf/pldi/BonaertDBV21}.
Given an input data $x$ and a transformer $F(x)$, transformer verification identifies a maximal bound $\epsilon$, such that all inputs $x'$ that are ``close'' to the input data (\textit{i.e.}, $|x'-x|\le \epsilon$) cannot ``mislead'' the transformer (\textit{i.e.}, $F(x) = F(x')$).
A larger $\epsilon$ indicates better robustness. 
Early verification approaches \cite{reluplex} enumerate all possible inputs $x'$ that satisfy $|x'-x|\le \epsilon$ and conduct inference on each input to check predictions.
These approaches show prohibitive latency due to the large number of inputs $x'$.
Recent transformer verification \cite{transformer_verification,transformer_verification1} avoids such enumeration by providing a single pair of lower and upper bounds for transformer predictions over all these inputs, as illustrated in \autoref{fig:intro_example}.
We can verify the robustness of a transformer if the lower bound of the correct prediction is higher than the upper bound of other predictions.
The key computing pattern is a \textit{bound-centric computation}, which computes a pair of inequality bounds for individual neurons.
It first represents the input perturbations with inequality bounds over input neurons (\textit{e.g.}, $x-\epsilon \le x' \le x+\epsilon$) and then propagates these bounds across layers to generate the bounds for transformer predictions.

\begin{figure}
    \centering
    \includegraphics[width=\linewidth]{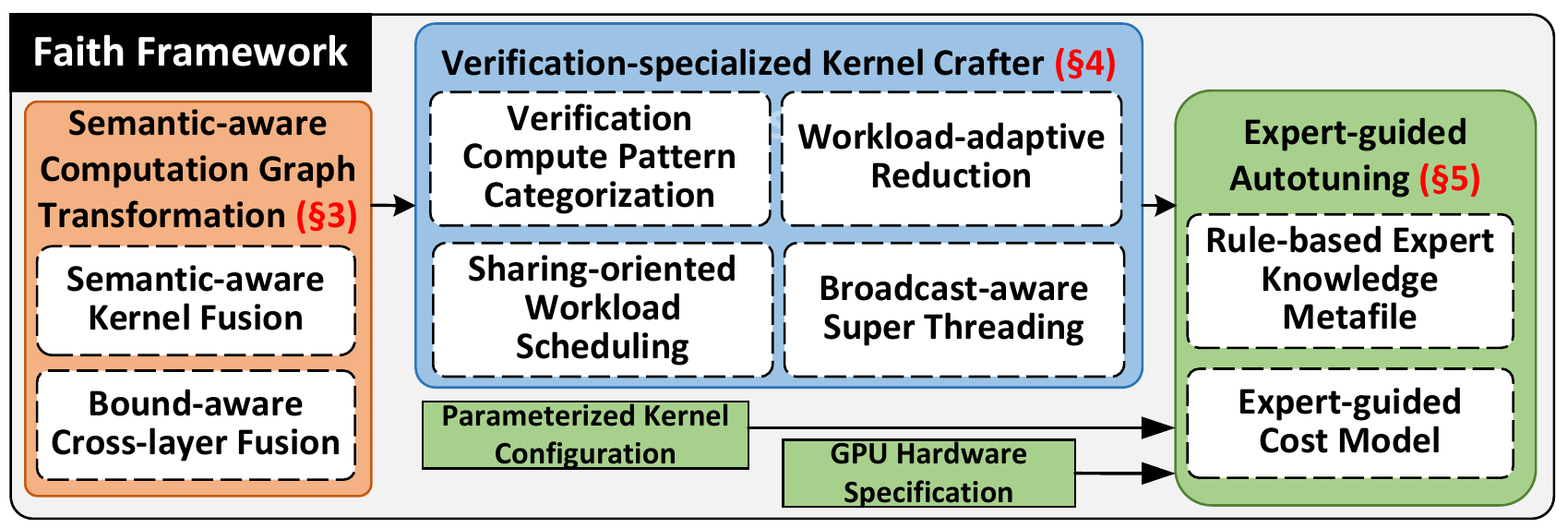}
    \vspace{-15pt}
    \caption{Overview of Faith Framework}
    \vspace{-8pt}
    \label{fig:overview}
\end{figure}


While transformer verification can formally verify the robustness of transformers, it also introduces high latency and limits its applications.
In particular, transformer verification usually leads to second-level latency \cite{transformer_verification} in contrast to millisecond-level latency of standard transformers.
We identify three challenges behind efficient transformer verification.




\textbf{Lack of performance optimization over transformer verification computing patterns.}
Existing transformer verifications usually utilize the existing deep learning (DL) frameworks, such as PyTorch \cite{PyTorch}, which are designed for standard NNs.
However, transformer verification shows significantly different computing patterns from standard NNs due to the nature of bound-centric computation.
For example, when computing the upper bound of an output neuron, transformer verification needs to use the upper bound of the input neuron if the weight is positive; and the lower bound of the input neuron if negative.
Straightforwardly deploying transformer verification to the existing DL frameworks usually leads to poor performance.

\textbf{Lack of framework support for verifying diverse NN layers.}
Transformer verification shows large diversity in the bound computation for different types of NN layers such as projection layer with only perturbed features and self-attention layer with both perturbed weights and features.
Even for the same type of NN layers, diverse upper bounds and lower bounds may be designed which requires different implementations.
For example, Crown \cite{crown} utilizes two ReLU bound designs for generating more precise bounds for verification, where these bounds are selected dynamically according to the range of input neurons.
This diversity makes it challenging to hand optimize GPU kernels in transformer verification.


\textbf{Lack of verification-specialized adaptability towards modern GPUs.}
Transformer verification involves abundant memory-intensive operations such as reduction and broadcast.
These memory-intensive operations can usually be significantly accelerated with rich architecture supports (\textit{e.g.}, warp-level synchronized reduction) in modern GPUs.
However, existing DL frameworks usually only focus on computation-intensive operations (\textit{e.g.}, convolution) and ignore abundant optimization opportunities for memory-intensive operations.
This leads to significant overhead in transformer verification with a large number of memory-intensive operations.

In this paper, we build \textbf{Faith}, the first framework for efficient transformer verification on GPUs.
We show an overview of the Faith framework in \autoref{fig:overview}.
First, we propose \textit{\textbf{semantic-aware computation graph transformation}} to 
fully exploit fusion opportunities in transformer verification at the computation graph level.
Our key insight is that transformer verification shows significantly different computing patterns (\textit{e.g.}, two kernels for computing lower and upper bounds involve similar input data) from standard NNs.
These computing patterns usually exhibit abundant data reuse opportunities.
By exploiting such semantic information, Faith can fully harvest performance potential in transformer verification and achieve significant speedup over existing DL frameworks.

Second, we propose a \textit{\textbf{verification-specialized kernel crafter}} to optimize transformer verification towards modern GPUs.
Transformer verification contains abundant memory-intensive operations, such as elementwise computation, reduction, and broadcast. These operations may have complex dependencies and lead to performance bottlenecks.
To this end, Faith automatically exploits a set of GPU architecture supports to improve the parallelism of such operations.
Moreover, Faith introduces a set of optimizations to effectively mitigate memory access and improve performance by exploiting GPU memory hierarchies.

Third, we propose \textit{\textbf{expert-guided autotuning}} to efficiently search optimized implementations in the large search space.
Existing DL frameworks~\cite{TVM,Ansor} usually conduct autotuning in a hardware-agnostic approach where an ML-based cost model is deployed to implicitly learn hardware impact over performance from scratch.
Instead, we propose a rule-based expert knowledge metafile to explicitly provide a small set of hardware characterizations and an expert-guided cost model to incorporate the expert knowledge.
Faith exploits these two components to achieve efficient schedule exploration in the large design space of transformer verification.

\begin{figure*}
    \centering
    \includegraphics[width=0.95\textwidth]{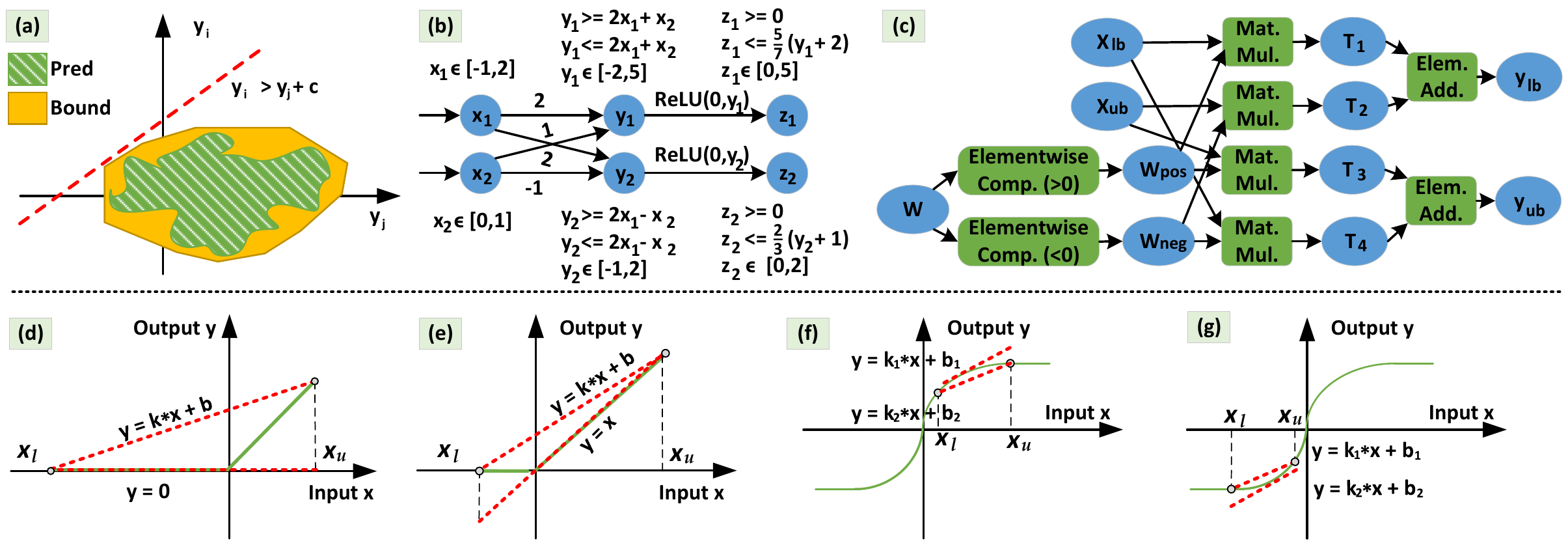}
    \vspace{-10pt}
    \caption{Illustration of transformer verification. (a) model prediction and verification bound; (b) an example of verifying a model with a fully connected layer and a ReLU layer; (c) computation graph of projection layer in transformer verification; (d)-(e) two types of bounds for ReLU layer; (f)-(g) two types of bounds for the \textit{Tanh} layer.} 
    \vspace{-10pt}
    \label{fig:background}
\end{figure*}

In summary, this paper makes the following contributions:
\vspace{-5pt}
\begin{itemize}
    \item We build Faith, the first efficient framework to optimize the performance of transformer verification on GPUs.
\vspace{-7pt}
    \item We propose a set of verification tailored system optimizations. In particular, we design a \textit{semantic-aware computation graph transformation} to identify and exploit novel fusion opportunities for transformer verification, a \textit{verifier-specialized kernel crafter} to effectively map transformer verification kernels to GPU backends, and an \textit{expert-guided autotuning} to incorporate a set of expert knowledge on modern GPU architecture to guide large design space exploration.
\vspace{-7pt}
    \item Extensive experiments show that Faith achieves up to $3.4\times$ speedup ($2.6\times$ on average)  over state-of-the-art frameworks. 
\end{itemize}
\vspace{-10pt}

%% file: text/2_Background.tex
\vspace{-5pt}
\section{Related Work and Motivation} \label{sec:background}
\vspace{-5pt}
In this section, we first introduce the background of transformer verification (\autoref{sec:verifier_background}).
Then, we discuss related work on DL frameworks (\autoref{sec:NNCompiler}).
Finally, we present opportunities and challenges for efficient transformer verification on GPUs (\autoref{sec:opportunity_challenge}).

\vspace{-8pt}
\subsection{Transformer Verification} \label{sec:verifier_background}
\vspace{-5pt}

\quad\textbf{Standard Transformers.}
Transformer \cite{transformer,BERT,xlnet,roberta} takes a sentence as input and predicts a label for this sentence (\textit{e.g.}, hate speech or benign speech).
Given a sentence with $Length$ tokens, we usually first map each token to a pretrained embedding \cite{word2vec} of dimension $Dim\_in$ and represent the feature of a sentence as a tensor of shape $Length \times Dim\_in$.
For a batch of sentences, we have input feature $X$ as a tensor of shape $Batch\_size \times Length \times Dim\_in$, where $Batch\_size$ is the number of sentences in a batch.
Since the number of tokens varies across sentences, $Length$ is set to the maximal number of tokens over all sentences in a batch.


A transformer has three types of operators.
The first type is the \textit{elementwise operator} that applies computation on individual feature scalars.
For example, on each scalar $x$ in the input feature, we have $ReLU(x) = max(0, x)$ and $Tanh(x) = \frac{e^{2x}-1}{e^{2x}+1}$.
The second type is the \textit{matrix multiplication operator} that takes an input tensor $X$, a weight matrix $W$, and generates an output tensor $Y=XW$.
We note that these two types are similar to operators in prior neural networks.
The third type is the \textit{dot product operator}, which is the key idea behind the transformer model.
Informally speaking, it takes two input tensors $Q$ and $K$ of the same shape $Batch\_size \times Length \times Dim\_in$.
Then, it computes an output tensor $Y= Q^TK$ of shape $Batch\_size \times Length \times Length$ to measure the pairwise similarity between individual words in a sentence.
This similarity can significantly improve the learning capacity of the model and the prediction accuracy.

\textbf{Adversarial Attack on Transformers.}
Adversarial attack \cite{DBLP:journals/corr/GoodfellowSS14,attack1,attack2,attack3,attack4,attack5} identifies small perturbations to input data $X$ that can change the transformer prediction.
Formally, consider a transformer $f(\cdot)$, an input sentence $X$, and a tolerable input perturbation bound $\epsilon$, where the transformer correctly classifies $X$ as a label $i$ (\textit{e.g.}, hate speech).
In other words, the sentence has label $i$ and $y_i > y_j$ for any $j\neq i$ where $y_i$ is the predicted probability.
Adversarial attack identifies a slightly perturbed sentence $X' = X+\eta$ such that $\eta \in B(0,\epsilon)$ and there exists a label $j$ (\textit{e.g.}, benign speech) such that $y_i < y_j$.
This perturbed sentence $X'$ is an \textit{adversarial example}.


\textbf{Transformer Verification.}
Transformer verification \cite{reluplex,transformer_verification,transformer_verification1,DBLP:conf/pldi/BonaertDBV21} computes a maximum bound $\epsilon$ and mathematically proves that there does not exist an adversarial example $X'$ within the $\epsilon$-ball of $X$ (\textit{i.e.}, $(X'-X) \in B(0, \epsilon)$).
Verifying transformers is challenging since transformers are essentially non-convex functions.
The key idea of transformer verification is to utilize linear bounds as an approximation to NN predictions.
We illustrate transformer verification at the model prediction layer in \autoref{fig:background}(a).
Given these linear bounds, transformer verification can simply check if the predictions insides the bounds satisfy certain linear requirements, such as $y_i > y_j + c$, where $c$ is a positive number.
As illustrated in \autoref{fig:background}(a), this bound-based approach is sound since the linear bound covers the non-convex area of NN predictions.

We show an example of bound-centric computation of transformer verification in \autoref{fig:background}(b).
Consider a fully connected layer $Y[j] = \sum_{i=1}^n W[j,i]\cdot X[i]$ where $Y[j]$, $W[j,i]$, and $X[i]$ are scalars.
Here, we skip the index for batch size and length for notation simplicity.
A formal summary of notations can be found  in \autoref{tab:notation}.
For each neuron $X[i]$, there is a lower and a upper bound
\begin{equation*} \small
    X[i] \ge X_{lb}[i] + X_{lw}[i]* \vec \epsilon, \;\;X[i] \le X_{ub}[i] + X_{uw}[i]* \vec \epsilon
\end{equation*}
where $X_{lb}[i]$ and $X_{ub}[i]$ are scalars, $X_{lw}[i]$, $X_{uw}[i]$, and $\vec \epsilon$ are vectors.
For the input neurons, we have $X_{lb}[i] = X_{ub}[i] = X[i]$, $X_{lw}[i]$ and $X_{uw}[i]$ are one-hot vectors with $1$ at the index $i$ and $0$ at other indices.
Given this linear bound, we can compute \textit{concretized} bounds for each neuron as
\begin{equation} \small \label{eq:concretization}
    X_l[i] = X_{lb}[i] - \epsilon * ||X_{lw}[i]||, \; \; X_u[i] = X_{ub}[i] + \epsilon * ||X_{uw}[i]||
\end{equation}
where $||\cdot ||$ computes the norm with reduction operations.

When computing the bounds for output neuron $Y[j]$, we note that bound computation depends on the sign of weights $W[j,i]$.
In particular, we have upper bounds $Y_{ub}[j]$ as 
\begin{equation} \small \label{eq:NNVerifier}
\begin{split}
    Y[j] \le & Y_{ub}[j] + Y_{uw}[j]* \vec \epsilon\\
    = & (\sum_{W[j,i]\ge 0} W[j,i]\cdot X_{ub}[i] + \sum_{W[j,i]<0} W[j,i]\cdot X_{lb}[i]) \\
    + & (\sum_{W[j,i]\ge 0} W[j,i]\cdot X_{uw}[i] + \sum_{W[j,i]<0} W[j,i]\cdot X_{lw}[i])) * \vec \epsilon 
\end{split}
\end{equation}
The lower bounds can be computed in a similar way.
This bound computation (\autoref{eq:NNVerifier}) is significantly different from standard NN computation since it explicitly considers the sign of weights.
Previous transformer verification directly exploits the standard DL frameworks to build a computation graph (\autoref{fig:background}(c)) for computing bounds, which leads to inefficient memory access and computation overhead.
We will discuss the opportunities and challenges of efficient transformer verification in \autoref{sec:opportunity_challenge}.


\begin{table}[t] \small
    \centering
    \caption{Notations in transformer verification.}
    \vspace{-5pt}
    \scalebox{0.9}{
    \begin{tabular}{c|l}
    \hline
    \hline
        $W$ & Transformer weights. Shape: $Dim\_in \times Dim\_out$\\
        $X$ & Input feature tensor. Shape: $Batch\_size \times Length \times Dim\_in$ \\
        \hline
        \multirow{2}{*}{$X_{lb}$, $X_{ub}$} & The tensor of lower and upper bound bias of input features. \\ 
                           & Shape: $Batch\_size \times Length \times Dim\_in$\\
        \multirow{2}{*}{$X_{lw}$, $X_{uw}$} & The tensor of lower and upper bound weights of input \\
                           & features. Shape: $Batch\_size \times Length \times Dim\_in \times Dim\_out$\\
        \multirow{2}{*}{$X_{l}$, $X_u$} & The tensor of concretized lower and upper bounds of input \\ 
                           & features. Shape: $Batch\_size \times Length \times Dim\_in$ \\
    \hline
    \hline
    \end{tabular}}
    \label{tab:notation}
    \vspace{-15pt}
\end{table}

For the same NN layer, diverse bound computation designs may still be developed to provide tighter bounds on NN predictions.
We illustrate two types of bounds for the ReLU layer in \autoref{sec:background}(d)-(e) and two types of bounds for the Tanh layer in \autoref{sec:background}(f)-(g).
A tighter bound (\textit{i.e.}, less space between linear bounds and ReLU function) is preferred to provide a better linear bound approximation to NN prediction.
For example, consider the concretized lower bound $X_l[i]$ and upper bound $X_u[i]$ for an input neuron $X[i]$, when we have $abs(X_l[i]) > abs(X_u[i])$, linear bound in \autoref{fig:background}(d) is preferred over the linear bound in \autoref{fig:background}(e) since the former one provides a tighter approximation.
This diversity in bound design adds more complexity to developing frameworks for transformer verification.



\vspace{-5pt}
\subsection{Deep Learning Frameworks on GPUs} \label{sec:NNCompiler}
\vspace{-5pt}

GPUs have been widely exploited to accelerate deep learning workload \cite{APNN-TC,QGTC,DBLP:conf/ppopp/WangYZWLSXK18,YiShi,DBLP:conf/icde/ZhangPZZSM021}.
Efficiently mapping deep learning workloads to the GPU computing and memory hierarchy is usually the key to improve performance \cite{GNNAdvisor,turbotransformer,Yishi2,DBLP:journals/vldb/0007ZSWCMC021,DBLP:journals/tpds/ZhangZSMD22}.
GPU computing hierarchy contains threads, warps, and blocks \cite{cuda-programming-guide}.
Each block has multiple warps and each warp has exactly $32$ threads that compute with single-instruction-multiple-data (SIMD).
GPU memory can be generally treated as a hierarchy of registers, shared memory, and global memory.
Accessing registers is much faster than accessing shared memory, which is faster than accessing global memory.
Each thread can only access its own registers and threads in a block cannot access shared memory from other blocks.

Many DL frameworks~\cite{PyTorch,TVM,Ansor} have been developed recently to efficiently support NN workload on GPUs.
Early works such as PyTorch \cite{PyTorch} take user-specified computation graphs for neural networks and maps towards hand-tuned kernels on backend platforms (\textit{e.g.}, GPUs).
However, this approach usually builds upon kernels developed for standard NNs and cannot efficiently support transformer verification computation.
Recent works, such as TVM \cite{TVM} and Ansor \cite{Ansor}, can automatically generate such backend kernels based on a set of heuristic rules on fusion and operator optimizations.
However, these heuristic rules are developed specifically for standard NNs.
Naively incorporating these rules into transformer verification may lead to unsatisfactory performance due to the significant difference in computation patterns.
For example, \autoref{fig:background}(c) shows the computation graph for utilizing the kernels of standard NNs on transformer verification.
This approach leads to heavy sparsity and redundant memory access.
In particular, only half of the elements in $W_{pos}$ and $W_{neg}$ are non-zero values, leading to $50\%$ sparsity.
To this end, we build Faith, the first framework for efficient transformer verification on GPUs.

\begin{figure}
    \centering
    \includegraphics[width=0.95\linewidth]{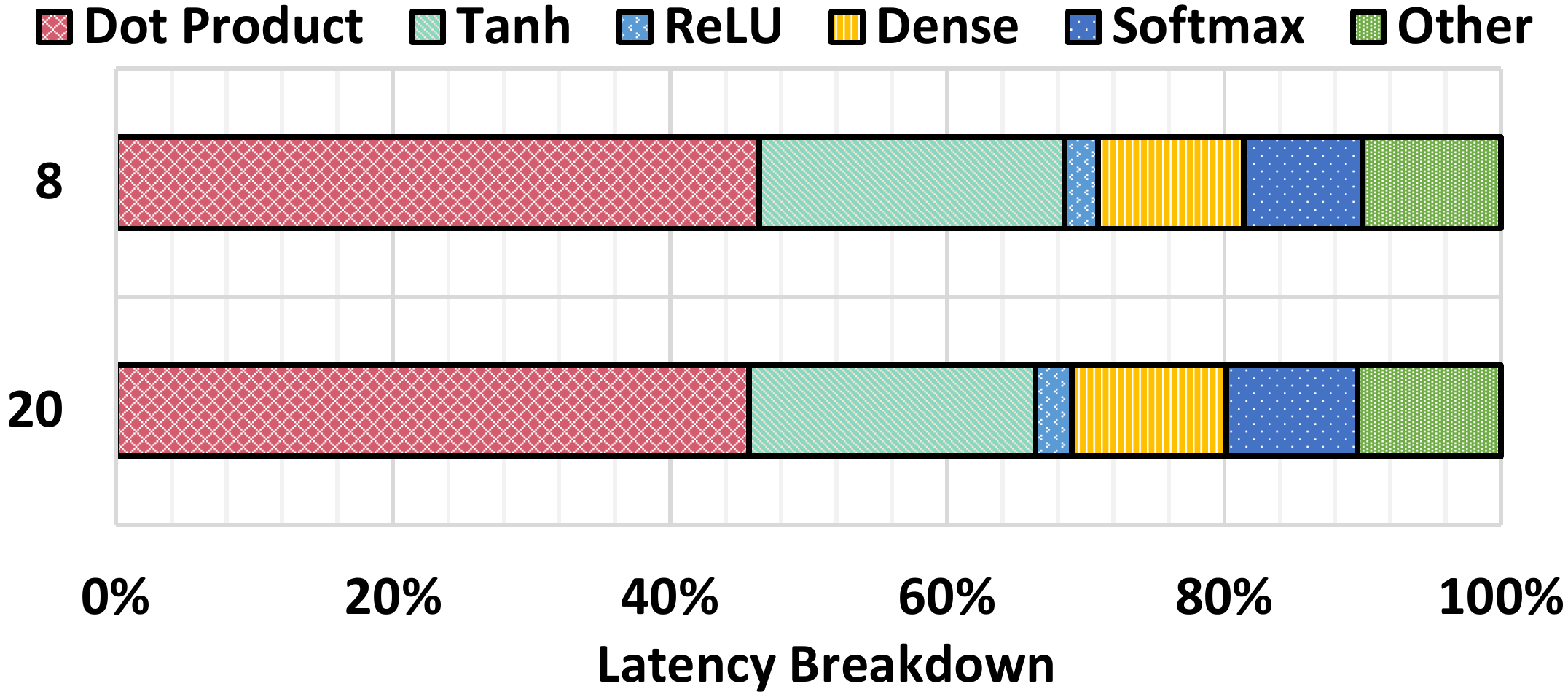}
    \vspace{-8pt}
    \caption{Latency breakdown of transformer verification on sentences with length $8$ and $20$. Here, we show the latency of verifying individual operators such as dot product and Tanh.} 
    \label{fig:breakdown}
    \vspace{-13.2pt}
\end{figure}

\vspace{-8pt}
\subsection{Opportunities and Challenges} \label{sec:opportunity_challenge}
\vspace{-5pt}
In this section, we introduce optimization opportunities and challenges in enabling efficient transformer verification.

We show the latency of verifying individual transformer operators in \autoref{fig:breakdown}.
We profile this latency breakdown based on the state-of-the-art transformer verification implemented with PyTorch \cite{PyTorch}.
We have three major observations.
First, dot product accounts for around $45\%$ latency.
Dot product takes two input tensors $Q$ and $K$ where both inputs may be perturbed during adversarial attack, which is significantly different from matrix multiplication that only one input (\textit{i.e.}, feature $X$) may be perturbed.
This adds complexity to the verification of dot product operators \cite{transformer_verification} and longer latency.
Second, elementwise operators such as Tanh and ReLU account for a large portion of latency in transformer verification.
This is significantly different from standard NNs where elementwise operators can usually be fused with remaining operators and show low latency.
Third, we observe that matrix multiplication and softmax accounts for certain latency.

\textbf{Opportunities:}
There are two major opportunities to accelerate transformer verification.
The first opportunity is to exploit the semantics of transformer verification to minimize redundant memory access and computation.
Our investigation shows that transformer verification has rich semantic information (\textit{e.g.}, $50\%$ sparsity in $W_{pos}$ and $W_{neg}$), which can be exploited to accelerate transformer verification.
The second opportunity is to exploit the modern GPU architectures to efficiently support diverse computing patterns in transformer verification.
One example is to accelerate abundant reduction computation in \autoref{eq:concretization}.


\textbf{Challenges:}
Although these ideas sound promising, the efforts to realize the benefits are non-trivial due to several challenges.
First, transformer verification shows significantly different computing patterns from standard NNs.
Straightforwardly borrowing optimizations for standard NNs such as kernel fusion can hardly bring similar benefits.
Second, while exploiting GPU architecture supports may bring benefits, we still need specialized designs as a synergy between architecture and specialized computing patterns.
Moreover, exploiting advanced GPU architecture supports will add more complexity to the search space of optimized kernels which motivates novel autotuning optimizations.

%% file: text/3_GraphOptimization.tex
\section{Semantic-aware Computation Graph Transformation} \label{sec:graphTransformation}
\vspace{-5pt}

In this section, we propose \textit{semantic-aware computation graph transformation} for efficient transformer verification.
We first propose \textbf{semantic-aware kernel fusion} to fuse kernels within a transformer layer.
It contains two novel types of fusions -- \textit{weight-paring based fusion} and \textit{double bound based fusion}.
Then, we propose \textbf{bound-aware cross-layer fusion} to efficiently fuse kernels across transformer layers.



\begin{figure}
    \centering
    \includegraphics[width=\linewidth]{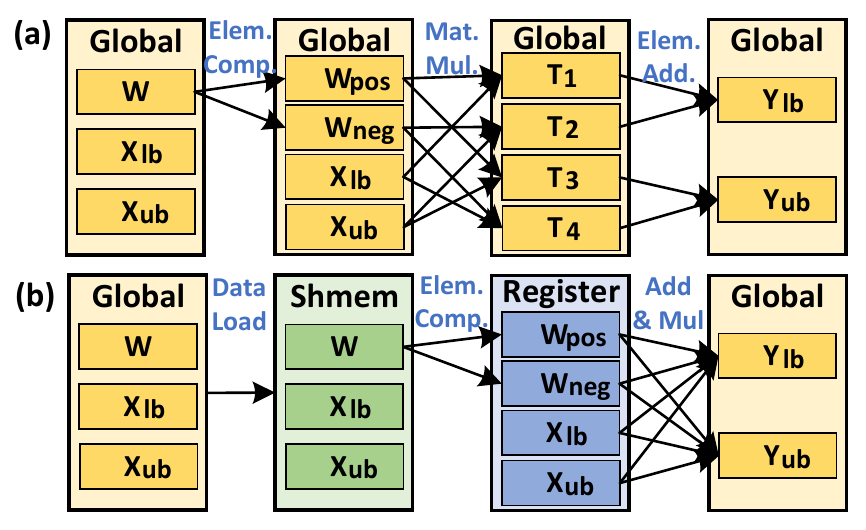}
    \caption{Illustration of Semantic-aware Kernel Fusion. We show the memory access pattern before and after applying semantic-aware kernel fusion in (a) and (b), respectively.}
    \label{fig:kernelFusion}
    \vspace{-6pt}
\end{figure}



\vspace{-5pt}
\subsection{Semantic-aware Kernel Fusion} \label{sec:semantic-aware-kernel-fusion}
\vspace{-2pt}

The semantic-aware kernel fusion fuses operators in a single transformer layer to minimize memory access.
Different from standard transformers, a single layer in transformer verification usually involves multiple kernels to compute the bounds adaptively to the sign of weights, as discussed in \autoref{sec:verifier_background}.
Existing transformer verification \cite{transformer_verification,transformer_verification1} usually uses a set of GPU kernels developed for standard transformers to serve the need for transformer verification.
We illustrate the memory access pattern of this baseline approach in \autoref{fig:kernelFusion}(a).
These kernels need to independently read data from the global memory of GPUs and lead to heavy memory overhead.
Moreover, these kernels fail to exploit semantic information in transformer verification and show heavy redundancy during memory access.
For example, baseline approaches usually first split the weight matrix $W$ into two weight matrices $W_{pos}$ and $W_{neg}$ according to weight signs and then use each matrix for computing lower and upper bounds.
Here, these two split matrices $W_{pos}$ and $W_{neg}$ have the same shape of $M\times N$ as the weight matrix $W$.
However, reading these matrices independently requires loading $2MN$ scalars, which leads to redundant memory access.


We propose semantic-aware kernel fusion to minimize such memory overhead by exploiting transformer verification semantics and GPU memory hierarchies (\textit{i.e.}, global memory, shared memory, and registers).
We illustrate our semantic-aware kernel fusion in \autoref{fig:kernelFusion}(b).
Our key insight is to first load data collaboratively from global memory and only distinguish data semantics (\textit{e.g.}, $W_{pos}$ and $W_{neg}$) at the register level to mitigate redundant memory access.
In particular, we identify \textit{weight-paring based fusion} and \textit{double bound based fusion} as the two most important semantics in transformer verification.

\textbf{Weight-pairing based fusion.}
We first propose weight-paring-based fusion to mitigate redundant memory access when reading $W_{pos}$ and $W_{neg}$.
Our key observation is that the zero values in $W_{pos}$ are exactly the position of non-zero values in $W_{neg}$.
Formally, we have $W_{pos}+W_{neg} = W$.
To this end, instead of using an operator to split weight matrix $W$ into $W_{pos}$ and $W_{neg}$, we first load the matrix $W$ from global memory to shared memory without distinguishing the sign of individual scalars.
Then, we split the weight matrix $W$ into $W_{pos}$ and $W_{neg}$ when loading data from shared memory to registers, as illustrated in \autoref{fig:kernelFusion}(b).
In our design, we only need to load $MN$ scalars from global memory, which leads to significantly reduced memory access compared with loading $2MN$ scalars in baseline approaches.




\textbf{Double bound based fusion.}
Our second optimization is a double-bound-based fusion.
One important semantics in transformer verification is to multiply the same weight matrix with lower and upper input bounds (\textit{e.g.}, $X_{lb}$ and $X_{ub}$) to compute the output bounds (\textit{e.g.}, $Y_{lb}$ and $Y_{ub}$ in \autoref{fig:kernelFusion}(b)).
Meanwhile, when computing the bound for output neurons, we usually need to read both lower and upper bounds for computation.
For example, when computing the upper bound of output neurons, we need to read upper bound when weight is positive and read lower bound when weight is negative.
Suppose the input bounds $X_{lb}$ and $X_{ub}$ have shape $N\times K$, we need to load $4NK$ scalars during transformer verification.

Instead, we propose to fuse the computation of lower and upper bounds such that the lower and upper bounds only need to be loaded once to save memory access.
In particular, we first use threads across GPU blocks to collaboratively load tiles of input matrices from global memory to shared memory, which can be accessed by different GPU threads.
Here, we use shared memory to enable data sharing across GPU threads since different threads may multiply the same input bound scalar with different weight scalars (\textit{e.g.}, multiplying the first row in $X_{lb}$ and $X_{ub}$ with various columns in $W$).
Then, each thread loads independent data from shared memory to registers and directly accumulates output bounds $Y_{lb}$ and $Y_{ub}$ in registers.
We note that this design further improves performance by eliminating the redundant global memory access during generating $Y_{lb}$ and $Y_{ub}$.




\vspace{-5pt}
\subsection{Bound-aware Cross-layer Kernel Fusion} \label{sec:cross-layer-kernel-fusion}

Bound-aware cross-layer kernel fusion fuses the verification of kernels across multiple transformer layers to further minimize memory access.
Existing frameworks for accelerating standard NNs usually rely on a set of rules to fuse kernels.
One popular example is to fuse convolution kernel with the following elementwise kernels (\textit{e.g.}, ReLU kernel for elementwise comparison with $0$).
However, these rules usually cannot be applied to fuse kernels for transformer verification.
For example, verifying the ReLU kernel requires first a concretization operation with a global reduction to compute the concretized bounds for a neuron and then applies different computation according to the concretized bounds (see \autoref{sec:verifier_background}).

To this end, we propose a set of rules for cross-layer kernel fusion in transformer verification.
In particular, we recognize three types of operators.
The first type is \textit{input-reduction-compute} that conducts reduction or concretization operation on the input data before computation.
One example is verifying nonlinear activation functions such as \textit{ReLU} and \textit{Tanh} that requires concretized bounds to apply different computation.
Another example is the \textit{softmax} operator that computes a global summation for normalization.
The second type is \textit{strict-elementwise} that contains only elementwise computation and does not require concretization or global summation.
The third type is \textit{dense-computation} such as matrix-matrix multiplication kernels.
In our cross-layer kernel fusion design, we can always fuse a \textit{dense} operator with its following \textit{strict-elementwise} operator.
However, we cannot fuse \textit{dense} operator with \textit{input-reduction-compute} due to the concretization or reduction operation.
In addition, we can fuse \textit{input-reduction-compute} with its following \textit{strict-elementwise} operator.
Finally, we can fuse multiple \textit{strict-elementwise} operators (\textit{e.g.}, elementwise addition and multiplication).






%% file: text/4_KernelOptimization.tex
\vspace{-2pt}
\section{Verification-specialized Kernel Crafter} \label{sec:kernelCrafter}
\vspace{-2pt}

In this section, we propose a verification-specialized kernel crafter to efficiently map transformer verification towards modern GPUs.
We exploit intrinsic properties (\textit{e.g.}, abundant reduction operations) of transformer verification which are significantly different from standard transformer operators.
One major challenge in building the kernel crafter is the large diversity in verification designs across operators (see \autoref{fig:background}(d)-(g)).
To tackle this challenge, we first propose a \textit{verification pattern categorization} to abstract such diversity and provide a small set of computing patterns over verification of diverse operators.
Then, we propose three optimizations to efficiently support these computing patterns of transformer verification.

\vspace{-2pt}
\subsection{Verification Pattern Categorization} \label{sec:compute_pattern}
\vspace{-2pt}

While there are diverse bound designs across different operators, we characterize transformer verification into four typical computing patterns.
Based on this characterization, Faith can abstract the diversity in bound designs into a combination of computing patterns and exploit optimizations towards individual computing patterns for improving performance.
Similar to standard NNs, one important computing pattern is \textit{generalized matrix multiplication (GEMM)} when verifying projection layers and fully connected layers.
Matrix multiplication is the major bottleneck in standard NNs and has been well-optimized by existing DL frameworks. 
Besides GEMM, transformer verification introduces three other time-consuming computing patterns, which are highlighted as follows:

The first computing pattern is \textit{generalized vector reduction}.
One typical source of generalized vector reduction is concretization that computes the norm and generates the concretized lower and upper bounds for individual neurons (see \autoref{eq:concretization}).
Formally, consider a matrix $X=[\vec x_1, \vec x_2, \cdots, \vec x_m] \in \mathbb{R}^{m\times n}$ where $\vec x_i = [x_{i,1}, x_{i,2}, \cdots, x_{i,n}]$ are vectors of length $n$.
The generalized vector reduction computes an output $Y = [y_1, y_2, \cdots, y_n] \in \mathbb{R}^n$ that satisfies
\begin{equation}  \small \label{eq:vector-reduction}
    y_i = reduction(\vec x_i) = \sum_{j=1}^n f(x_{i,j}), \;\; i \in \{1,2,\cdots ,m\}
\end{equation}
Here, $f(x)$ is an elementwise function that takes a scalar input and generates a scalar output.
One example for $f(x)$ is $x^2$ when computing the $L_2$ norm for input vectors.

The second computing pattern is \textit{generalized elementwise multiplication} which appears frequently when verifying elementwise operators such as ReLU and Tanh.
Formally, consider a concretized lower bound $l \in \mathbb{R}^{m\times n}$ and an upper bound $u \in \mathbb{R}^{m\times n}$ where $l_{i,j}$ and $u_{i,j}$ are concretized lower and upper bounds for the neuron at position $(i,j)$.
Let $X \in \mathbb{R}^{m\times n}$ be the input values.
The generalized elementwise multiplication computes an output $Y \in \mathbb{R}^{m \times n}$ that satisfies
\begin{equation} \small \label{eq:elementwise-multiplication}
    y_{i,j} = f(l_{i,j}, u_{i,j}) * x_{i,j}, \;\;\; i \in \{1,2,\cdots m\}, j\in \{1,2,\cdots, n\}
\end{equation}
Here, transformer verification introduces a function $f(\cdot, \cdot)$ that takes the lower and upper bounds for an input neuron and computes a scaling parameter which is multiplied with the input value of this neuron.
One example is the tangent line between the concretized lower and upper bounds when verifying Tanh layer, which accounts for more than 20\% latency as we profiled in \autoref{fig:breakdown}.
Another example is $f(l_{i,j}, u_{i,j}) = 1$ when verifying ReLU layer and $l_{i,j}$ is non-negative.
While $f(\cdot, \cdot)$ shows large diversity across operators, we stress that the same computing pattern is shared across these operators such that a uniform framework can be applied to improve performance.

The third computing pattern is \textit{generalized scalar-vector multiplication}.
This computing pattern exists widely when verifying dot products in the self-attention layer of transformers.
This computing pattern accounts for more than 40\% latency in transformer verification, as discussed in \autoref{fig:breakdown}.
Formally, consider a vector $S = [s_1, s_2, \cdots, s_m] \in \mathbb{R}^{m}$ and a matrix $X = [\vec x_1, \vec x_2, \cdots, \vec x_m] \in \mathbb{R}^{m\times n}$, where $s_i$ are scalars and $\vec x_i = [x_{i,1}, x_{i,2}, \cdots, x_{i,n}]$ are vectors of length $n$.
The generalized scalar-vector multiplication computes an output $Y = [\vec y_1, \vec y_2, \cdots, \vec y_n] \in \mathbb{R}^{n\times n}$ that satisfies
\begin{equation} \small \label{eq:scalar-vector}
\begin{split}
    \vec y_i = f(s_i) * \vec x_i = [f(s_i) *x_{i,1}, f(s_i) *x_{i,2}, \cdots, f(s_i) *x_{i,n}], \;\; \\ 
    i \in \{1,2,\cdots,m\}
\end{split}
\end{equation}
Here, $f(\cdot)$ is a function that takes a scalar input and generates a scalar output.

\textbf{Generability to diverse NN operators.}
Faith can effectively support verifying diverse NN operators such as SiLU and Leaky ReLU.
Our key insight is that verifying diverse NN operators usually share the same generalized computing pattern while the concrete computation formula might be different.
For example, $SiLU(x) = \frac{x}{1+e^{-x}}$ is an activation function that has significantly different concrete computation formula from $ReLU(x) = max(0,x)$.
However, both verifying SiLU and ReLU can be treated as the generalized elementwise multiplication (\autoref{eq:elementwise-multiplication}) and the same optimizations can be applied to improve performance.



In the following sections, we first demonstrate a \textit{workload-adaptive reduction} to improve the performance of generalized vector reduction (\autoref{eq:vector-reduction}). We then propose a \textit{sharing-oriented workload scheduling} to improve the performance of generalized elementwise multiplication (\autoref{eq:elementwise-multiplication}).
Finally, we demonstrate \textit{broadcast-aware super threading} to efficiently support the generalized scalar-vector multiplication (\autoref{eq:scalar-vector}).


\begin{figure}
    \centering
    \includegraphics[width=\linewidth]{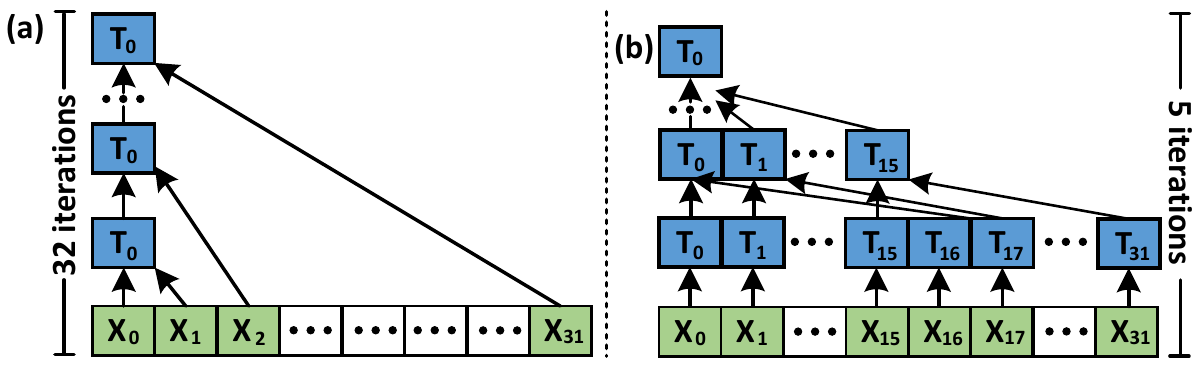}
    \vspace{-15pt}
    \caption{Illustration of Workload-adaptive Reduction. (a) Sequential Mode; (b) Parallel Mode. Here, $x_i$ and $T_i$ are the i-th data and thread, respectively.}
    \label{fig:reduction}
    \vspace{-10pt}
\end{figure}

\vspace{-5pt}
\subsection{Workload-adaptive Reduction} \label{sec:reduction}
\vspace{-5pt}

Transformer verification contains abundant reduction operations where a sequence of scalars are summed up into one scalar.
One common reduction operation is the concretization operation that computes the concretized lower and upper bounds for individual neurons, as discussed in \autoref{sec:background}.
Another common reduction operation is the softmax operation that is applied in each self-attention layer for measuring the relationship between individual words.
These reduction operations pose challenges between parallelism and data locality.
One baseline approach is to use a single thread to read and accumulate a sequence of scalars as illustrated in \autoref{fig:reduction}(a).
However, this approach usually leads to low parallelism and fails to exploit abundant threads in GPUs.
For example, we need $32$ iterations to accumulate $32$ scalars.
Another baseline approach is to first split this sequence of scalars into multiple chunks and allocate one thread to each chunk for accumulation.
Then, each thread writes the accumulated results for each chunk to global memory and uses an additional thread to finally accumulate the sum of each chunk.
While this approach improves parallelism, it requires expensive global memory access and high overhead.

\textbf{Workload-adaptive Reduction with length $n=32$.}
We propose a \textit{workload-adaptive reduction} to fully exploit GPU memory hierarchies and the inter-register communication functionalities.
We illustrate our design in \autoref{fig:reduction}(b).
Our design achieves high parallelism by enabling multiple threads for reduction simultaneously.
Meanwhile, we avoid the expensive data communication through global memory and exploit only efficient registers.
In particular, we use $32$ threads (\textit{i.e.}, a warp) to read these $32$ scalars simultaneously from global memory.
Considering these $32$ scalars are consecutive in global memory, we can efficiently load them with $32$ threads through coalesced memory access.
Then, we exploit the specialized instruction \texttt{\_shfl\_down\_sync}  to directly communicate data in registers across individual threads.
As illustrated in the parallel mode of \autoref{fig:reduction}(b), our design involves only five iterations of cross-thread data communication to generate the final accumulated result, rather than the $32$ iterations in the sequential mode of \autoref{fig:reduction}(a).

\textbf{Workload-adaptive Reduction with Arbitrary Length $n$.}
For an arbitrary length $n$, one naive approach is to repeatedly use $32$ threads to reduce $32$ scalars and then use $1$ thread to accumulate the final results.
However, this approach may lead to unnecessary communication across threads.
Suppose we are accumulating a vector of length $n = 32k$, we need $5$ iterations for reducing every $32$ scalars, leading to $5k$ iterations in total for accumulating the vector.
Instead, we propose a \textit{hybrid mode} to minimize the number of iterations while still achieving high parallelism.
In particular, we first split the input sequence into chunks where each chunk contains 32 scalars.
Then, we use $32$ threads to read one chunk simultaneously from global memory and  accumulate individual chunks iteratively.
For example, the 1-st thread accumulates the 1-st scalar in each chunk.
Here, the accumulation is conducted in registers and does not require communication across threads.
Finally, we apply a single $5$-iteration reduction across $32$ threads.
In total, our design has only $k+5$ iterations which are significantly less than $6k$ iterations in the naive approach.

\vspace{-5pt}
\subsection{Sharing-oriented Workload Scheduling}
\vspace{-5pt}

We propose \textit{sharing-oriented workload scheduling} to efficiently verify elementwise operators.
Different from standard transformers, verifying elementwise operators, especially non-linear ones (\textit{e.g.}, ReLU and Tanh), accounts for a large portion of latency in transformer verification as we discussed in \autoref{fig:breakdown}.
Verifying these operators usually first requires computing a concretized lower bound $X_l$ and upper bound $X_u$ for each input neuron and then computes the bounds for the output neuron.
Different signs of concretized input bounds usually lead to different computations for output bounds, which could easily lead to warp divergence and unsatisfactory performance.
Moreover, when computing the output bound weights (\textit{i.e.}, $Y_{lw}$ and $Y_{uw}$) for a neuron, we need to repeatedly use the same input bounds which leads to extra memory overhead.
\begin{figure}
    \centering
    \includegraphics[width=\linewidth]{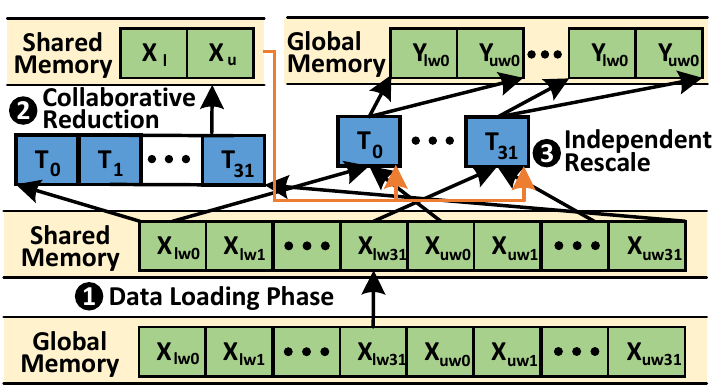}
    \vspace{-5pt}
    \caption{Illustration of sharing-oriented workload scheduling}
    \label{fig:sharing}
    \vspace{-10pt}
\end{figure}

To efficiently verify elementwise operators, we propose \textit{sharing oriented workload scheduling} to minimize memory access and improve performance.
Our key observation is that the same set of input bound weights $X_{lw}$ and $X_{uw}$ are used to compute the concretized input bounds $X_{l}$ and $X_{u}$, while these input weights are also used for computing the output bound weights $Y_{lw}$ and $Y_{uw}$.
Instead of repeatedly loading $X_{lw}$ and $X_{uw}$, we can exploit the GPU memory hierarchies to cache $X_{lw}$ and $X_{uw}$ and minimize the global memory access to improve the overall performance.

As illustrated in \autoref{fig:sharing}, we use a set of $T$(=32) threads to first (Step \circled{1}) load input bound weights $X_{lw}$ and $X_{uw}$ from global memory to shared memory.
Here, $T$ is a hyper-parameter to balance the parallelism and compute intensity, which will be selected in \autoref{sec:autotuning}.
Then (Step \circled{2}), these $T$ threads load input bound weights from shared memory and collaboratively compute the concretized lower and upper bounds $X_l$ and $X_u$, following our design in \autoref{sec:reduction}.
These concretized lower and upper bounds are stored in shared memory which can be accessed by individual threads.
Finally (Step \circled{3}), each thread independently loads individual $X_{lw}$ and $X_{uw}$ scalars from shared memory and rescales according to the concretized bounds $X_l$ and $X_u$.
Here, all threads in a warp are computing the output bound weights for the same neuron and the concretized input bounds are the same across threads in a warp.
Thus, all threads in a warp can apply the same rescaling computation and avoid warp divergence.
We also note that input bound weights are only loaded once from global memory which mitigates redundant global memory access.

\vspace{-5pt}
\subsection{Broadcast-aware Super Threading} \label{sec:super-threading}
\vspace{-5pt}

We propose \textit{broadcast-aware super threading} to efficiently support generalized scalar-vector multiplication, as discussed in \autoref{eq:scalar-vector}.
One naive approach is to use one thread to read a scalar $s_i$ and a vector $\vec x_i$ and computes the generalized scalar vector multiplication $f(s_i) \vec x_i$.
However, this approach fails to exploit the parallelism opportunities in generalized scalar vector multiplication.
Another approach is to split the vector $\vec x_i$ into multiple chunks and use one thread for each chunk.
However, this approach requires threads to repeatedly read the same scalar $s_i$ from global memory and shows redundant memory access.

Instead, we propose a broadcast-aware super threading to achieve high parallelism while minimizing memory access.
We consider two types of super threading for generalized scalar vector multiplication.
The first type is a group of $32$ threads (\textit{i.e.}, a warp for one vector).
When using $32$ threads to compute the multiplication between a scalar $s_i$ and a vector $\vec x_i$, these $32$ threads can read the scalar $s_i$ once, broadcast across threads with modern GPU memory, and compute $f(s_i)$ simultaneously.
Based on this broadcast, we can mitigate the redundant memory access that each thread repeatedly read the same scalar $s_i$.
The second type is a group of $32t$ threads (\textit{i.e.}, $t$ warps for one vector).
In this case, we use one warp to read the scalar $s_i$ and use shared memory to broadcast $s_i$ across warps.




%% file: text/5_ExpertGuidedAutotuning.tex
\vspace{-10pt}
\section{Expert-guided Autotuning Optimization} \label{sec:autotuning}
\vspace{-5pt}
Considering the large design space of optimization towards GPUs, one natural question arises: \textit{Can we effectively incorporate hardware knowledge to find optimal operator implementation?}

Existing works such as TVM \cite{TVM} and Ansor \cite{Ansor} usually autotune operator implementations in a hardware-agnostic way.
In particular, these works extract implementation-specific parameters such as tiling size and use a cost model to implicitly learn the relationship between these parameters and performance.
However, there are two drawbacks in this hardware-agnostic approach.
First, there is a complex interaction between implementation and the hardware properties, which could be hard to be implicitly learned by the cost model.
For example, existing works \cite{DBLP:conf/ppopp/FengWCZ0D21,DBLP:conf/ppopp/YanWC20,DBLP:conf/cgo/LaiS13,DBLP:conf/ppopp/Li0YJL19} on hand-tuning large matrix-matrix multiplication operators usually maximize the number of registers in use to improve cache performance.
However, this optimization is also limited by the number of registers for each GPU thread since exceeding such limitation may lead to register spilling \cite{register_spilling} and a significant performance drop.
A careful reasoning on the interaction between the implementation-specific parameters (\textit{e.g.}, the number of registers for caching data) and the hardware properties (\textit{e.g.}, the number of registers per thread) is usually necessary to maximize the performance.
To tackle this challenge, we propose an \textit{expert-guided autotuning optimization} to automatically reason both implementation-specific parameters and hardware properties.
In particular, we have the following designs.

\textbf{Rule-based Expert Knowledge Metafile.}
We propose a \textit{rule-based expert knowledge metafile} to capture hardware properties.
This metafile only needs to be set once for each type of GPUs and requires limited manual efforts.
In particular, we consider two types of rules.
The first type is \textit{hard rules} which represents hardware limitation such as the maximal shared memory size and the maximal number of registers per thread.
Violating these rules may lead to significant performance drop such as register spilling.
The second type is \textit{soft rules} which represents intrinsic trade-offs related to the hardware properties such as the number of streaming multiprocessors (SM) and the number of threads per SM.
One typical design choice is the number of threads per block which will be mapped to threads on the same SM.
Allocating more threads per block usually leads to better parallelism for the sub-task assigned to a block.
However, allocating more threads per block may also hinder executing multiple blocks on the same GPU SM hardware and lead to worse overall parallelism.

\textbf{Expert-guided Cost Model.}
We propose an \textit{expert-guided cost model} to automatically tackle the complex interaction between implementation-specific parameters and hardware properties.
Given a set of candidate operator implementations, we have two phases to select the optimal implementation.
The first phase is to estimate the shared memory and register usage for each candidate. We rule out candidates that consume more shared memory and registers than hardware capacity, as specified in the expert knowledge metafile. 

The second phase is to train a cost model for the remaining candidates and use the cost model to select the best candidate. We use XGBoost \cite{xgboost} as the cost model. It takes as input the implementation-specific parameters (e.g., tiling size) for candidates and the hardware properties (e.g., shared memory size). We use the cost model to predict the latency of candidates and select top-k candidates with low latency. Finally, we profile the latency of these top-k candidates on GPUs and use the profiled latency to further fintune our cost model.
We repeat this procedure for a pre-defined iterations (=5 by default) and select the implementation with the lowest latency.

When training the cost model, we construct the training dataset as follows. We randomly select a small number of candidates and use their implementation-specific parameters and the hardware properties as feature $X$. Then, we profile the latency of each candidate implementation on GPUs as the label $Y$. We collect these $(X, Y)$ as the training dataset to train the cost model.

%% file: text/6_Evaluation.tex
\begin{figure*}
\centering
\begin{subfigure}{.475\textwidth}
 \centering
 \includegraphics[width=\linewidth]{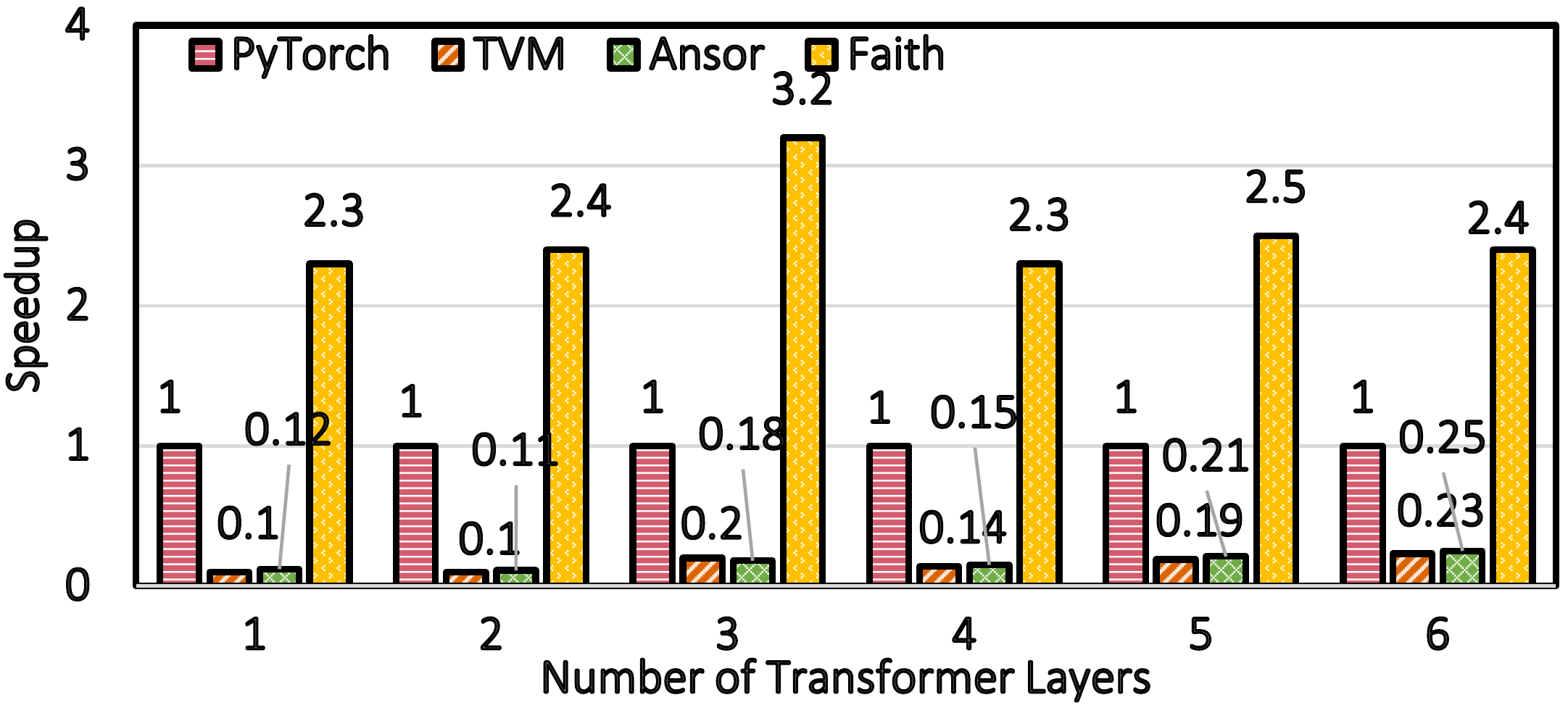}
\vspace{-15pt}
  \caption{On A100 GPU.}
\vspace{-10pt}
\end{subfigure}
\hfill
\begin{subfigure}{.475\textwidth}
 \centering
 \includegraphics[width=\linewidth]{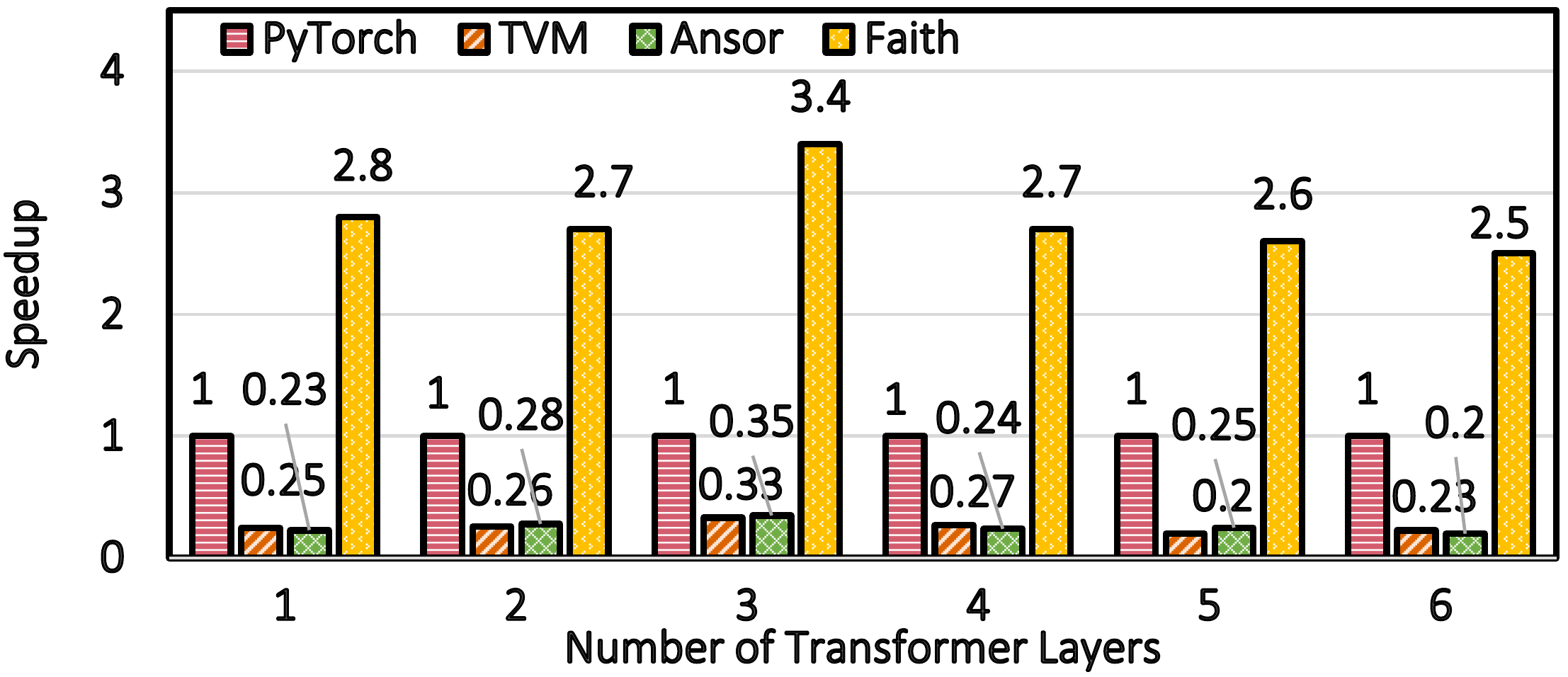}
    \vspace{-15pt}
    \caption{On V100 GPU.}
    \vspace{-10pt}
\end{subfigure}
\caption{Overall speedup on SST dataset.}
\vspace{-10pt}
\label{fig:overall-SST}
\end{figure*}

\begin{figure*}[t]
\centering
\begin{subfigure}{.475\textwidth}
 \centering
 \includegraphics[width=\linewidth]{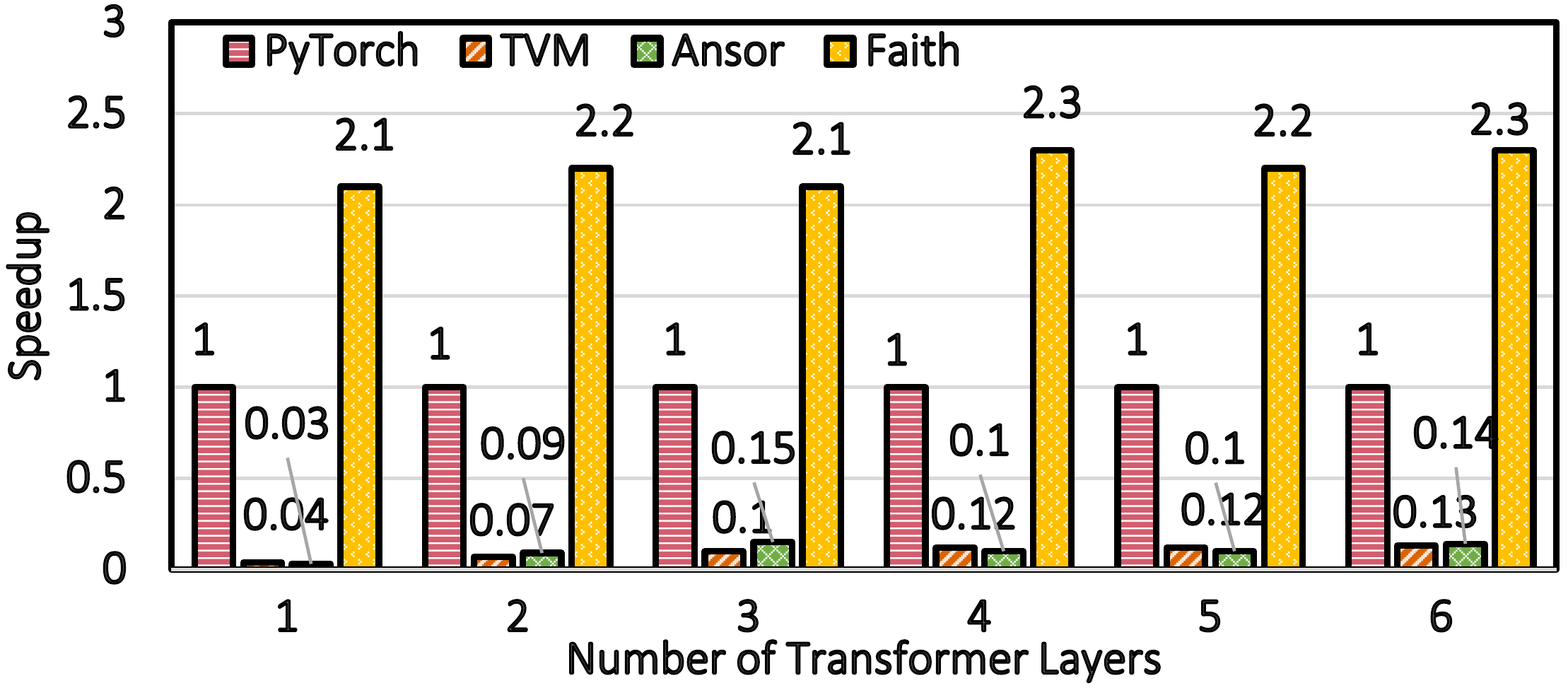}
\vspace{-15pt}
  \caption{On A100 GPU.}
\vspace{-10pt}
\end{subfigure}
\hfill
\begin{subfigure}{.475\textwidth}
 \centering
 \includegraphics[width=\linewidth]{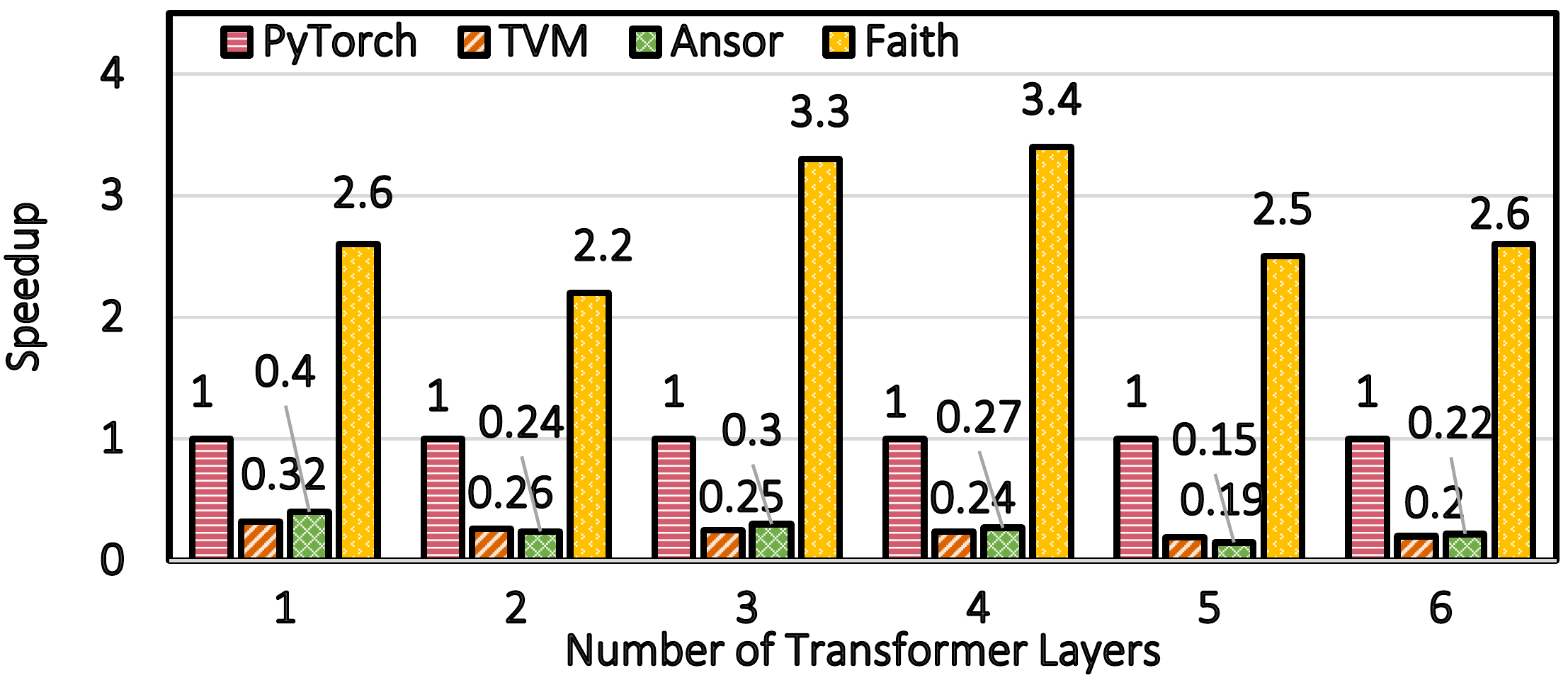}
    \vspace{-15pt}
    \caption{On V100 GPU.}
    \vspace{-10pt}
\end{subfigure}
\caption{Overall speedup on Yelp dataset.}
\label{fig:overall-Yelp}
\vspace{-10pt}
\end{figure*}

\begin{table}[t] \small
    \centering
    \begin{tabular}{c|c|c|c|c|c|c}
        \hline
        \multirow{2}{*}{\textbf{Dataset}} & \multirow{2}{*}{\textbf{\#Train}} & \multirow{2}{*}{\textbf{\#Val}} & \multirow{2}{*}{\textbf{\#Test}} & \multicolumn{3}{c}{\textbf{Length}}\\
        \cline{5-7}
        &&&& \textbf{min} & \textbf{mean} & \textbf{max}\\
        \hline \hline
        \textbf{SST} & 67,349 & 872 & 1,821 & 4 & 25 & 62 \\
        \textbf{YELP} & 560,000 & 0 & 38,000 & 5 & 98 & 128\\
         \hline
    \end{tabular}
    \vspace{-5pt}
    \caption{Dataset statistics}
    \label{tab:dataset}
    \vspace{-15pt}
\end{table}

\vspace{-5pt}
\section{Evaluation}
\vspace{-5pt}

In this section, we comprehensively evaluate Faith over various datasets and GPU backends.
We first present our experiment setup in \autoref{sec:setup}.
Then, we show the overall speedup on end-to-end transformer verification in \autoref{sec:e2e-evaluation}.
Finally, we provide more optimization analysis on individual transformer layers in \autoref{sec:optimization-analysis}.

\vspace{-5pt}
\subsection{Experiment Setup} \label{sec:setup}
\vspace{-5pt}

\textbf{Baselines.} We compare Faith with the state-of-the-art transformer verification \cite{transformer_verification} based on PyTorch.
We further compare with TVM \cite{TVM} and Ansor \cite{Ansor}, as stronger baselines.
TVM and Ansor are two state-of-the-art deep learning compilers for standard neural networks.
We feed the pytorch model into TVM and Ansor through relay frontend \cite{relay} which will automatically optimize transformer verification performance.
While TVM and Ansor take minutes to compile an operator implementation, we do not incorporate this compilation latency and record only inference latency for a fair comparison.

\textbf{Datasets.}
We evaluate two popular datasets, Yelp \cite{yelp} and SST \cite{sst}, following the setting in state-of-the-art transformer verification \cite{transformer_verification}.
These two datasets are widely used in the natural language processing for analyzing sentiment in languages.
We summarize the statistics of these two datasets in \autoref{tab:dataset}.
SST dataset contains 67,349 training sentences, 872 validation sentences, and 1,821 testing sentences.
In SST dataset, there are $4$ to $62$ tokens in each sentence and the average number of tokens in a sentence is $25$.
YELP dataset contains 560,000 sentences as training data and 38,000 sentences as testing data.
In YELP dataset, there are $5$ to $128$ tokens in each sentence and the average number of tokens in a sentence is $98$.

\textbf{Transformer Networks.} 
We evaluate Faith on transformer networks with $1$ to $6$ layers to demonstrate the performance on large models.
Following popular transformer settings, each transformer layer has 4 attention heads and an embedding size of 128.
Furthermore, we study the Faith performance under diverse embedding sizes in \autoref{sec:optimization-analysis}.

\textbf{Experiment Configuration.} 
We evaluate with an NVIDIA A100 GPU and an NVIDIA V100 GPU to show Faith performance on various GPU backends. 
The host server with A100 GPUs is an AMD EPYC 7742 64-Core Processor 
and runs Ubuntu 20.04 with CUDA 11.3.
The host server with V100 GPUs has 32 cores of Intel(R) Xeon(R) CPU E5-2620 v4 @ 2.10GHz 
and runs Ubuntu 16.04 with CUDA 10.1.

\vspace{-5pt}
\subsection{Overall Performance} \label{sec:e2e-evaluation}
\vspace{-5pt}
We show the overall speedup on SST dataset and Yelp dataset in \autoref{fig:overall-SST} and \autoref{fig:overall-Yelp}, respectively.
We show the performance improvement over transformers with diverse numbers of layers from $1$ to $6$, which covers popular settings in the natural language processing domain.
While the length of input sentences may have an impact on the performance improvement, we show the averaged speedup over all testing sentences in this section and study the impact of sentence length in \autoref{sec:optimization-analysis}.
We compare Faith with the PyTorch baseline following existing transformer verification open-source implementations \cite{transformer_verification}.
We further compare Faith with two state-of-the-art deep learning frameworks (\textit{i.e.}, TVM and Ansor) to provide a comprehensive comparison, as we discussed in \autoref{sec:setup}.

We show the overall speedup on SST dataset and A100 GPU in \autoref{fig:overall-SST}(a).
Compared with PyTorch, we observe $2.3\times$ to $3.2\times$ speedup ($2.5\times$ on average).
We contribute this performance improvement to our semantic-aware computation graph transformation (\autoref{sec:graphTransformation}) and verification-specialized kernel crafter (\autoref{sec:kernelCrafter}).
We further observe $17.2\times$ and $15.9\times$ speedup over TVM and Ansor, respectively.
The main reason is that TVM and Ansor focus on optimizing standard neural networks and fail to efficiently support verification-specific computing patterns, as discussed in \autoref{sec:NNCompiler}.
While Faith and these three baselines show different performance, we stress that the same verification bounds are generated, and the only difference resides in system optimizations.
Comparing across different numbers of transformer layers from $1$ to $6$, the performance improvement remains similar around $2.5\times$.
This result shows that Faith can efficiently support transformer verification with diverse numbers of transformer layers.
We show the overall speedup on SST dataset and V100 GPU in \autoref{fig:overall-SST}(b).
We have similar observation about the results on A100 GPU which shows that Faith can effectively adapt to diverse GPU backends, thanks to expert-guided autotuning optimization (\autoref{sec:autotuning}).

We show overall speedup on Yelp dataset and A100 GPU in \autoref{fig:overall-Yelp}(a).
Sentences in YELP dataset has $5$ to $128$ tokens ($98$ on average), which is longer than sentences in SST dataset with $4$ to $62$ tokens ($25$ on average).
This provides an opportunity to show Faith performance on long sentences.
Overall, we observe $2.1\times$ to $2.3\times$ speedup ($2.2\times$ on average) when comparing with the PyTorch baseline.
We also observe $26.7\times$ and $28.3\times$ speedup on average over TVM and Ansor, respectively.
This speedup is similar to the performance improvement on SST dataset and shows the good generality of Faith over diverse input data.
We also have similar observations on Yelp dataset and V100 GPU in \autoref{fig:overall-Yelp}(b).

\begin{figure*}
\centering
\begin{minipage}{.47\textwidth}
 \centering
 \includegraphics[width=\linewidth]{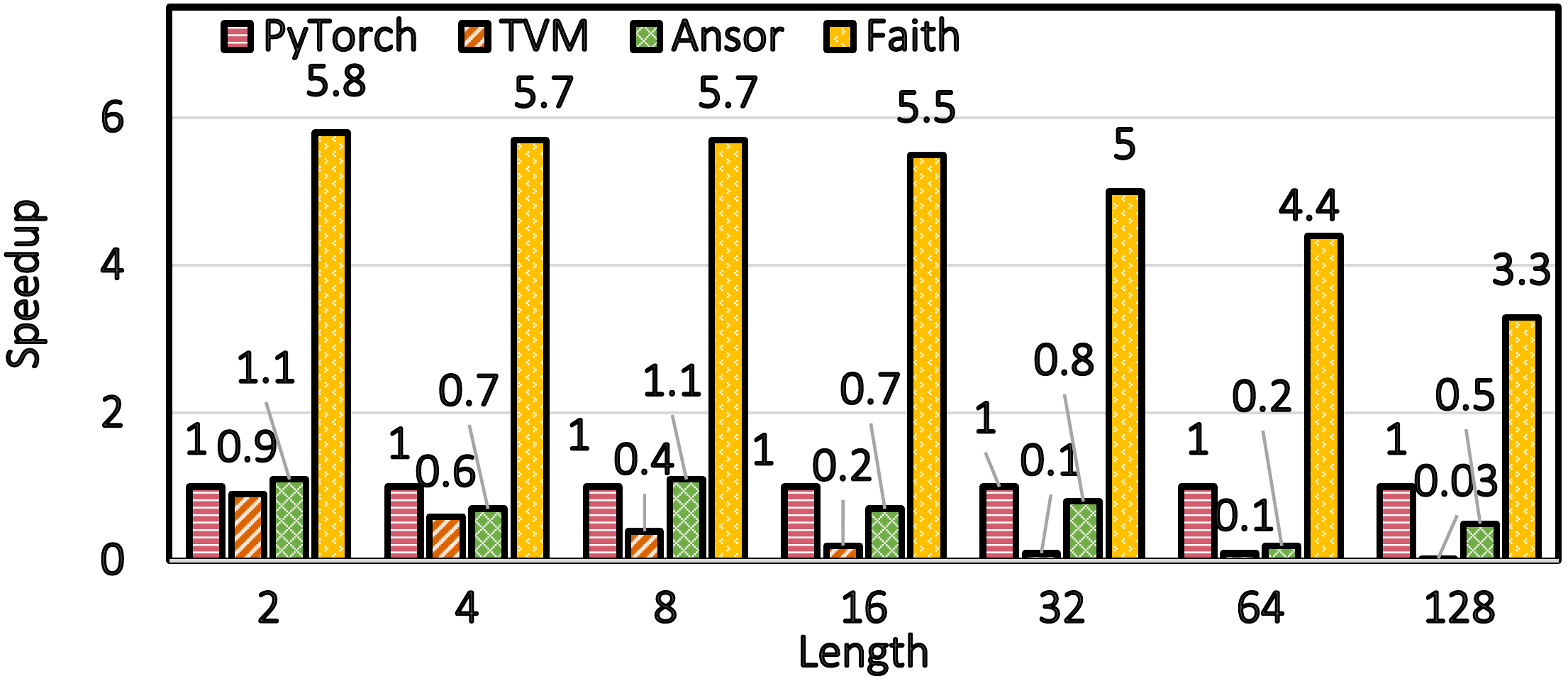}
 \vspace{-17pt}
  \caption{Speedup on verification of matrix multiplication over the diverse lengths.  Embedding Size: 128.} 
  \label{fig:matmul_length}
    \vspace{3pt}
\end{minipage}%
\hfill
\begin{minipage}{.47\textwidth}
 \centering
 \includegraphics[width=\linewidth]{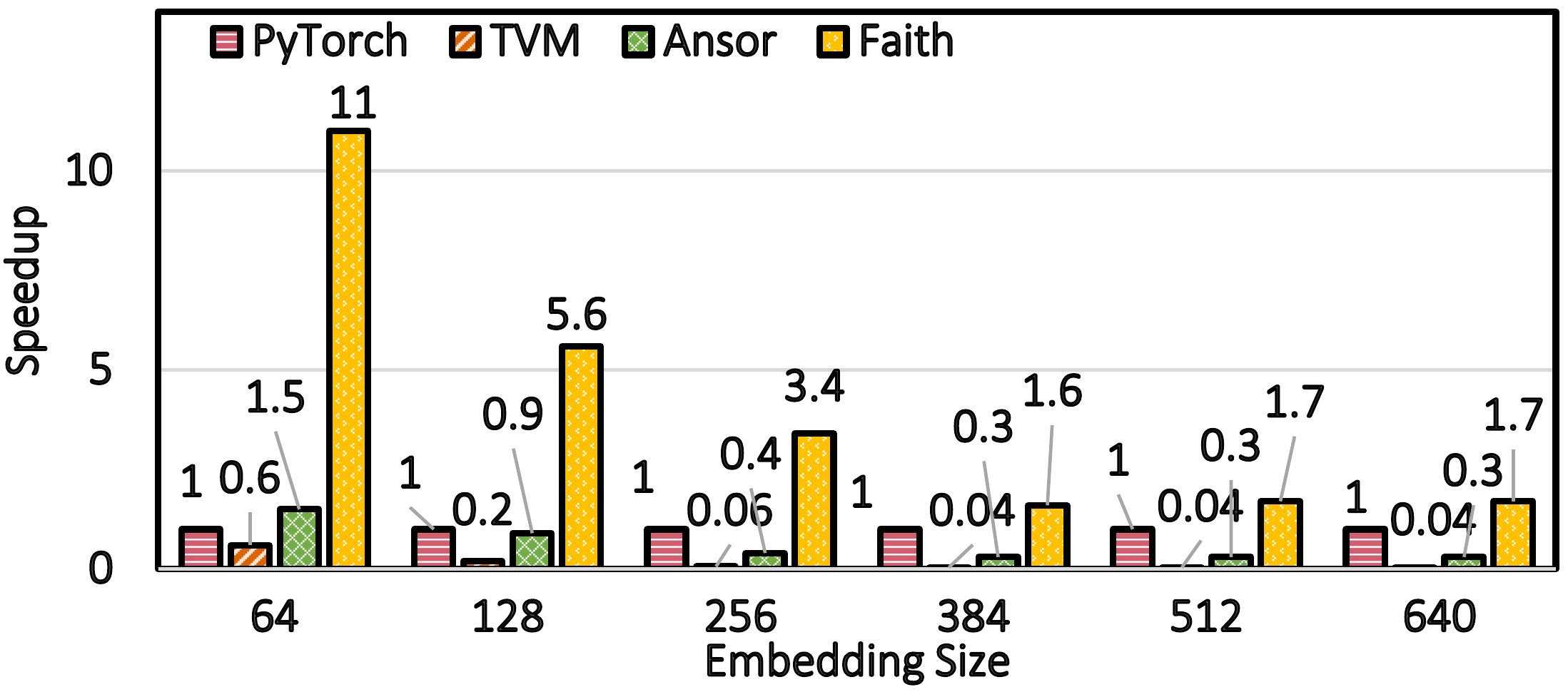}
 \vspace{-17pt}
  \caption{Speedup on verification of matrix multiplication over the diverse embedding sizes. Length: 16.}
  \label{fig:matmul_embedding_size}
    \vspace{3pt}
\end{minipage}

\begin{minipage}{.47\textwidth}
  \centering
  \includegraphics[width=\linewidth]{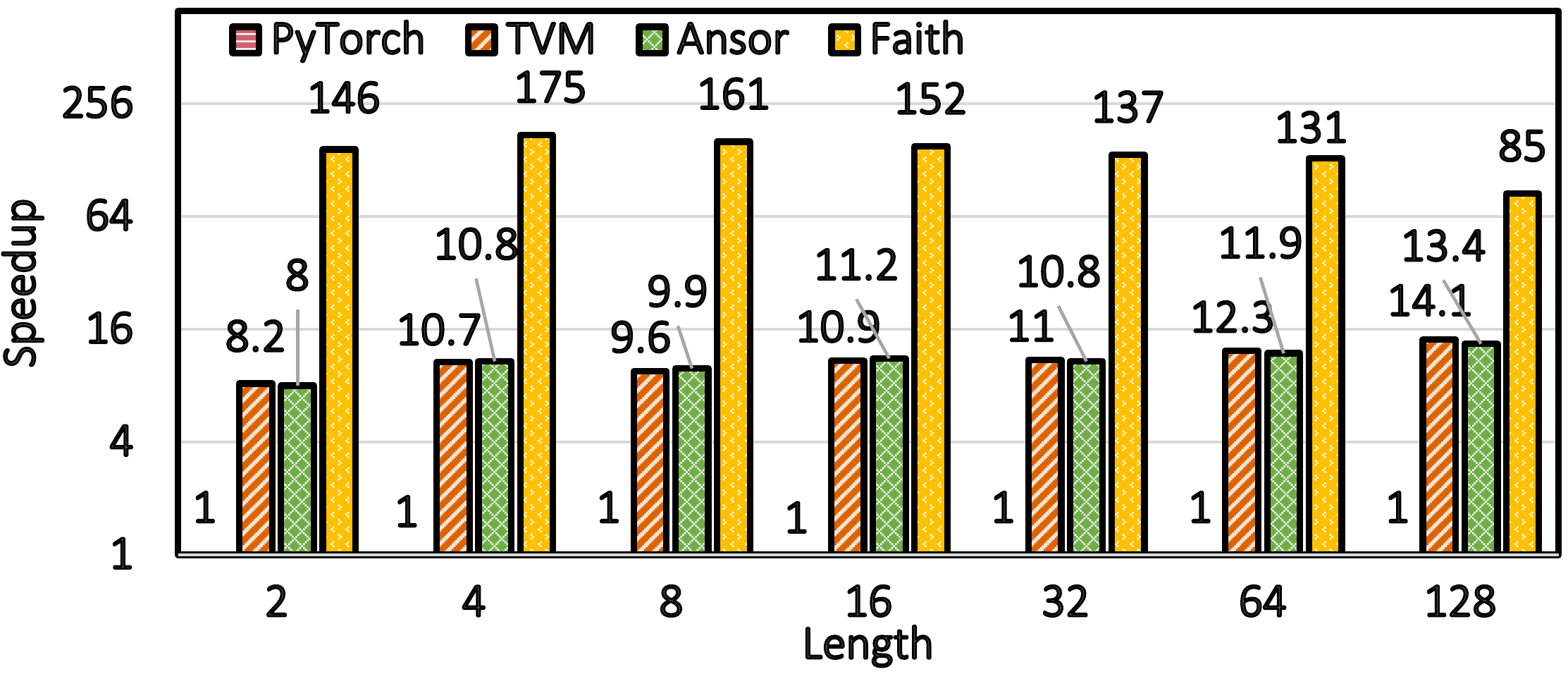}
  \vspace{-17pt}
  \caption{Speedup on verification of ReLU over the diverse lengths. Embedding Size: 128.}
  \vspace{-10pt}
  \label{fig:relu_length}
\end{minipage}
\hfill
\begin{minipage}{.47\textwidth}
  \centering
  \includegraphics[width=\linewidth]{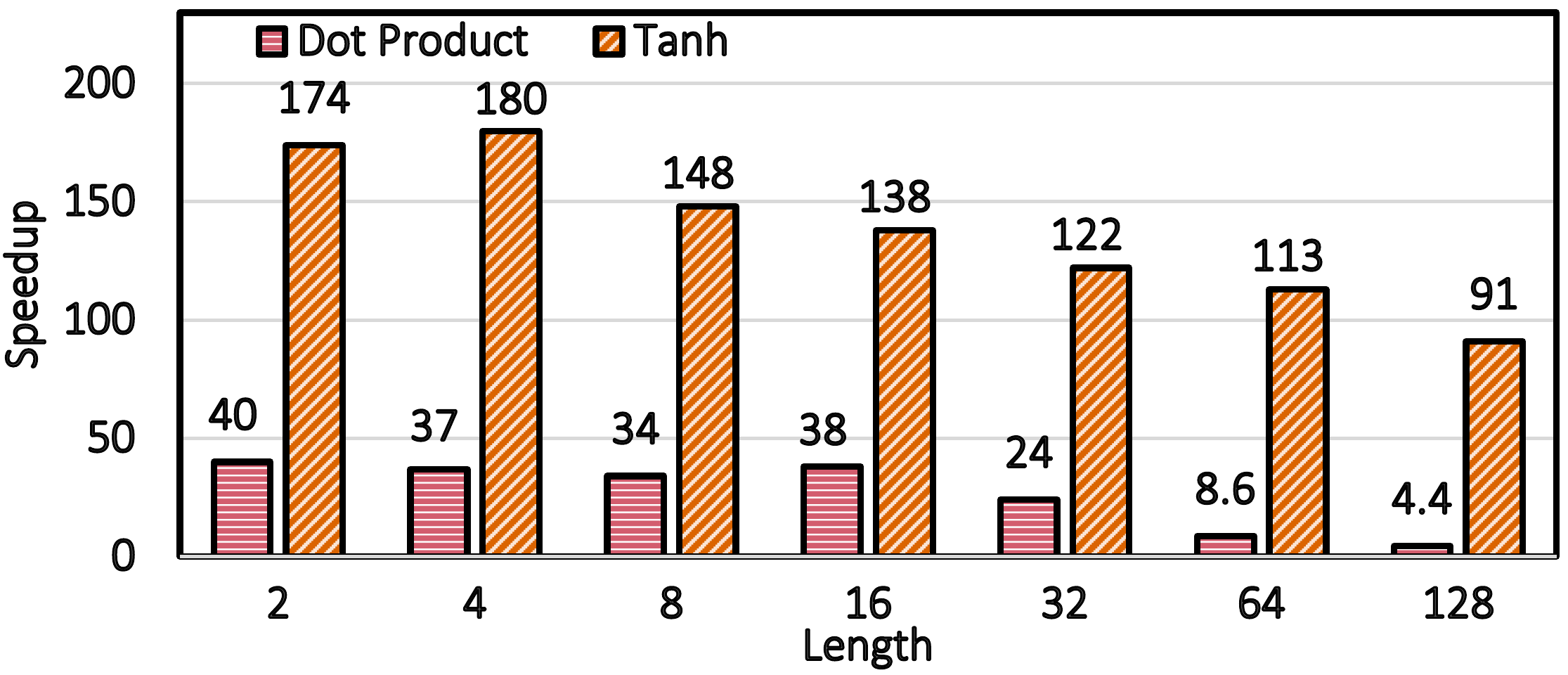}
  \vspace{-17pt}
  \caption{Speedup on verification of Tanh and dot product over the diverse lengths. Embedding Size: 128.}
  \vspace{-10pt}
  \label{fig:tanh_dotProduct_length}
\end{minipage}
\end{figure*}

\vspace{-5pt}
\subsection{Optimization Analysis} \label{sec:optimization-analysis}
\vspace{-5pt}

In this section, we show speedup from individual Faith optimizations.
We first show speedup on \textit{verification of matrix multiplication} over the diverse lengths and diverse embedding sizes.
Verification of matrix multiplication plays an important role in verifying projection layers and fully connected layers in transformers.
Then, we show the benefits on verification of ReLU, verification of dot product, and verification of Tanh, which in total accounts for around $70\%$ latency in transformer verification.
Since we observe similar performance on A100 GPU and V100 GPU, we focus on A100 GPU and omit results on V100 GPU in this section due to page limits.

\textbf{Performance benefits on verification of matrix multiplication.}
We show speedup on verification of matrix multiplication over the diverse lengths in \autoref{fig:matmul_length}.
We study the speedup over diverse lengths from $2$ to $128$, following the setting in the popular natural language processing datasets as summarized in \autoref{tab:dataset}.
Overall, we observe $5.1\times$ speedup on average over the PyTorch baseline.
This result shows significant performance benefits from utilizing Faith on accelerating transformer verification.
Comparing across lengths, we observe a higher speedup of $5.54\times$ over the PyTorch baseline on shorter sentences with $2$ to $32$ words.
The reason is that our autotuning optimization (\autoref{sec:autotuning}) automatically adjusts the number of threads and memory layout to improve the parallelism.
We achieve a smaller speedup of $3.85\times$ on longer sentences with $64$ and $128$ words.
For these longer sentences, we have achieved high occupancy on GPUs and the speedup is limited by the hardware capability.

Surprisingly, we observe that TVM and Ansor achieve $0.33\times$ and $0.73\times$ speedup, which is significantly slower than PyTorch baselines on verification of matrix multiplication.
The main reason is that TVM and Ansor focus on accelerating standard NNs and cannot efficiently support computing patterns in the verification of matrix multiplication (\autoref{fig:background}(c)).
Instead, Faith exploits a semantic-aware kernel fusion (\autoref{sec:semantic-aware-kernel-fusion}) to efficiently support such computing patterns in verification.


We show speedup on verification of matrix multiplication over the diverse embedding sizes in \autoref{fig:matmul_embedding_size}.
We study embedding size from $64$ to $640$ following popular transformer settings.
We note that transformer in natural language processing usually adopts a relatively small embedding size (\textit{e.g.}, $64$ to $256$), which is different from convolutional neural networks in computer vision that adopts a large embedding size (\textit{e.g.}, $1024$).
Overall, Faith achieves $4.2\times$ speedup on average over the PyTorch baseline.
This result shows that Faith can improve performance over diverse embedding sizes.
We also observe that Faith achieves larger speedup for smaller embedding sizes, which is similar to the case when verifying matrix multiplication over diverse lengths.

\textbf{Performance benefits on verification of ReLU.}
We show speedup on verification of ReLU over diverse lengths in \autoref{fig:relu_length}.
As we discussed earlier in \autoref{sec:compute_pattern}, verification of ReLU represents an important computing pattern of verifying elementwise operators.
Due to similar behaviors between diverse lengths and embedding sizes, we focus on verification over diverse lengths and keep embedding size as $128$, which is a popular setting in transformers.
Overall, Faith achieves $141\times$ speedup over PyTorch baseline.
This large speedup shows it promising to accelerate verification of elementwise operators.
Besides, Faith achieves $13.4\times$ and $13.5\times$ speedup over TVM and Ansor.
The reason is that our \textit{workload-adaptive reduction} (\autoref{sec:reduction}) can significantly improve parallelism during reduction and \textit{sharing-oriented workload sharing} can minimize memory access with GPU memory hierarchy.

\textbf{Performance benefits on verifying  Tanh and dot product layers.}
We show the speedup from Faith over the PyTorch baseline on verification of Tanh and verification of dot product in \autoref{fig:tanh_dotProduct_length}.
We skip the results of TVM and Ansor since these two frameworks do not support computing patterns in verification of Tanh and verification of dot product.
Here, we show results of verification of Tanh since it is a popular elementwise operator in transformer verification.
We also show results of verification of dot product since it accounts for around $45\%$ latency in transformer verification.
Overall, we observe that Faith achieves $138\times$ speedup on average for verification of Tanh.
This result is similar to the performance improvement for verification of ReLU, since both Tanh and ReLU are elementwise operators and share benefits from the same set of optimizations.
We also observe that Faith achieves $26.5\times$ speedup on average for verification of dot product.
This result shows the performance benefits from semantic-aware kernel fusion (\autoref{sec:semantic-aware-kernel-fusion}) and broadcast-aware super threading (\autoref{sec:super-threading}) that mitigate redundant memory access.

\begin{table}[]
    \centering
    \begin{tabular}{c|c|c|c|c|c|c}
        \hline 
        \textbf{\#Layers} & \textbf{1} & \textbf{2} & \textbf{3} & \textbf{4} & \textbf{5} & \textbf{6} \\
        \hline
        \hline
        \textbf{PyTorch} & 9.1 & 18 & 25 & 28 & 31 & 37 \\ 
        \hline
        \textbf{Faith} & 4 & 7.2 & 7.8 & 10.9 & 12.6 & 15.4 \\ 
        \hline
    \end{tabular}
    \vspace{-5pt}
    \caption{Latency on SST dataset and A100. Unit: Second.}
    \label{tab:raw-latency}
    \vspace{-15pt}
\end{table}

\textbf{Raw latency on transformer verification.}
We show the raw latency for transformer verification on the SST dataset and NVIDIA A100 GPU in \autoref{tab:raw-latency}. 
Faith requires only a few seconds to verify the NN prediction on a long sentence (with on average 25 tokens).
More specifically, when verifying transformers with $1$ to $6$ layers, Faith only requires $4$ to $15.4$ seconds to verifying a sentence.
This results brings transformer verification to the level of being practical for use.

%% file: text/7_Discussion.tex
\vspace{-5pt}
\section{Discussion}
\vspace{-5pt}

\textbf{Why Faith performs better than prior approaches.}
Existing frameworks, such as PyTorch, TVM, and Ansor, only support limited computation patterns for standard NNs. They cannot directly support bound-centric computation patterns in transformer verification.
While several frameworks like TVM allow autotuning for diverse operators, there is no magic. They still rely on hand-written GPU kernels (e.g., matrix multiplication) as the parametric templates (e.g., with tiling size as a parameter) and can only tune these tiling sizes. When applying to bound-centric computation patterns, they will break an operator for transformer verification into several hand-written GPU kernels for standard NNs. This leads to significantly higher memory access when aggregating computation results across GPU kernels into one transformer verification output.

Instead, Faith provides direct support for bound-centric computation patterns. Instead of breaking into several GPU kernels for standard NNs, we consider the bound-centric computation patterns as a whole and design a set of optimizations to reduce the memory and computation cost. For example, we found the lower and upper bounds are usually multiplied with the same weight matrix and can be loaded once to reduce memory overhead.

\textbf{Practicality of transformer verification with Faith.}
Faith brings transformer verification to be practical by consuming only around 10 seconds to verify a long sentence (e.g., 25 tokens).
We remark that transformer verification is one of the hottest topics in deep learning. Hundreds of related papers have been published in top deep learning conferences. The performance is essential to bring transformer verification into practical applications. However, existing efforts mainly reside in the algorithmic domain. In this paper, we build the first framework for efficient transformer verification on GPUs. Our work will open a new system research direction on developing high-performance systems for deep learning verification.

%% file: text/8_Concolusion.tex
\vspace{-5pt}
\section{Conclusion}
\vspace{-5pt}
Verifying the robustness of transformers draws increasing attention from both the academic and industry fields over the recent years.
Unfortunately, an efficient system design for transformer verification is still yet to come.
Existing transformer verification still exploits standard neural network frameworks which are unoptimized towards transformer verification workload.
The main reason is that efficient systems for transformer verification require both expertise from the machine learning community on mathematical verification designs and the system community on efficient memory and parallelism designs.

In this paper, we propose a Faith framework for efficient transformer verification.
Specifically, we first design a set of semantic-aware computation graph transformations to fully exploit fusion opportunities in transformer verification at the computation graph level.
Then, we propose a verifier-specialized kernel crafter to efficiently map fused verification kernels towards modern GPUs with minimized memory overhead and improved parallelism.
Finally, we propose an expert-guided autotuning to dynamically optimize kernels according to the transformer verification workload and GPU backend characteristics.
Comprehensive experimental evaluation shows that Faith significantly improves the performance of transformer verification over state-of-the-art frameworks.

Looking ahead, we believe our work in transformer verification would highlight a new direction on developing high-performance systems for deep learning verification.
This will encourage system experts with diverse backgrounds to build the next-generation deep learning systems and facilitate the wide application of secure deep learning.